%% file: iclr2026_conference.tex
\documentclass{article} 
\usepackage{iclr2026_conference_arxiv,times}

\input{math_commands.tex}

\usepackage{hyperref}
\usepackage{url}
\usepackage{booktabs}
\usepackage{xcolor}
\usepackage{graphicx}
\usepackage{subcaption} 
\usepackage{enumitem}
\usepackage{multirow}
\usepackage{array}
\usepackage{float}
\usepackage{makecell}
\usepackage[table]{xcolor}
\usepackage{wrapfig}
\usepackage{mwe}
\usepackage{pgfplots}

\usepackage{pifont}

\usepackage{tikz}
\usepackage{helvet}
\usetikzlibrary{calc,decorations.pathreplacing,arrows.meta}
\definecolor{inkc}{HTML}{000000}
\definecolor{mutedc}{HTML}{4D4D4D}
\definecolor{zsred}{HTML}{B2182B}
\definecolor{oursblue}{HTML}{1A5FA8}
\definecolor{regbg}{HTML}{B2182B}
\definecolor{zdpgreen}{HTML}{1B5E3A}
\newlength{\stripw}
\newcommand{\fC}{\sffamily\fontsize{5.8}{6.8}\selectfont}
\newcommand{\fD}{\sffamily\fontsize{6.6}{7.6}\selectfont}

\newcommand{\xmark}{\ding{55}}%

\newcommand{\eg}{\textit{e.g.}\@ }
\newcolumntype{H}{>{\setbox0=\hbox\bgroup}c<{\egroup}@{}}

\title{Disparity Has a Sign: Stereo Matching Beyond the Zero-Disparity Plane}

\author{%
  \begin{minipage}{\dimexpr\textwidth-2\tabcolsep\relax}\centering\normalfont
  ~\\[9pt]
  {\bf Jian Shi}$^{1}$ \quad {\bf Xinge Yang}$^{1}$ \quad {\bf Chaoyang Wang}$^{2}$ \quad
  {\bf Wolfgang Heidrich}$^{1}$ \quad {\bf Peter Wonka}$^{1}$ \\[9pt]
  $^{1}$KAUST, Thuwal, Saudi Arabia \qquad $^{2}$Independent Researcher \\[3pt]
  \url{https://shijianjian.github.io/ZDPShift/}
  \end{minipage}%
}

\iclrarxivcopy

\begin{document}

\maketitle

\begin{figure}[h]  
\centering
\renewcommand{\arraystretch}{0}
\small
\setlength{\stripw}{.925\linewidth}
\begin{tikzpicture}[inner sep=0pt,outer sep=0pt]
\node[anchor=north west] (S) at (0,0)
      {\input{secs/teaser_grid}};



\def\PH{0.7}\def\YM{19}
\coordinate (P0) at ($(S.south west)+(0,-0.7cm-\PH cm)$);
\coordinate (P1) at ($(S.south east)+(0,-0.7cm-\PH cm)$);
\foreach \i/\xx in {1/0.0996,2/0.2998,3/0.4999,4/0.7001,5/0.9004}{
  \coordinate (B\i) at ($(P0)!\xx!(P1)$);
}

\fill[regbg!7] ($(B2)!0.5!(B3)$) rectangle ($(P1)+(0,\PH cm)$);
\draw[line width=.4pt,regbg!35,dash pattern=on 1.4pt off 1.2pt]
      ($(B2)!0.5!(B3)$) -- ($(B2)!0.5!(B3)+(0,\PH cm)$);
\node[anchor=north east,font=\fC,text=regbg!85]
      at ($(P1)+(-2pt,\PH cm+19pt)$) {negative disparity regime};

\foreach \e in {0,10,20,30}{
  \draw[line width=.25,mutedc!20]
       ($(P0)+(0,\e/\YM*\PH cm)$) -- ($(P1)+(0,\e/\YM*\PH cm)$);
  \node[anchor=east,font=\fC,text=mutedc]
       at ($(P0)+(-2pt,\e/\YM*\PH cm)$) {\e};
}
\draw[line width=.5pt,mutedc] (P0) -- (P1);
\node[rotate=90,anchor=center,font=\fC,text=mutedc] at ($(P0)+(-14pt,0.5*\PH cm)$) {EPE (px)};

\foreach \i/\xx/\zs/\ou in {1/0.0996/2.2/1.9, 2/0.2998/0.6/0.8,
                            3/0.4999/7.40/2.0, 4/0.7001/13.9/1.5,
                            5/0.9004/29.3/1.4}{
  \coordinate (Z\i) at ($(P0)!\xx!(P1)+(0,\zs/\YM*\PH cm)$);
  \coordinate (O\i) at ($(P0)!\xx!(P1)+(0,\ou/\YM*\PH cm)$);
}
\fill[zsred,opacity=.10] (Z1)--(Z2)--(Z3)--(Z4)--(Z5)--(B5)--(B1)--cycle;
\draw[line width=1.1pt,zsred]    (Z1)--(Z2)--(Z3)--(Z4)--(Z5);
\draw[line width=1.1pt,oursblue] (O1)--(O2)--(O3)--(O4)--(O5);
\foreach \i in {1,...,5}{
  \fill[white] (Z\i) circle (1.9pt); \fill[zsred] (Z\i) circle (1.3pt);
  \fill[white] (O\i) circle (1.9pt); \fill[oursblue] (O\i) circle (1.3pt);
}
\draw[decorate,decoration={brace,mirror,amplitude=2.4pt},line width=.5pt,inkc]
     ($(O5)+(5pt,0)$) -- ($(Z5)+(5pt,0)$);
\node[anchor=west,font=\fD,text=inkc] at ($(O5)!0.5!(Z5)+(8.5pt,0)$)
     {$\mathbf{20.9\times}$};
\node[anchor=south east,font=\fC,text=zsred]    at ($(Z4)+(0,5pt)$) {SOTA prediction};
\node[anchor=north east,font=\fC,text=oursblue] at ($(O4)+(0,5pt)$) {\textbf{ours}};

\foreach \i/\lab in {1/{$-16$},2/{$0$},3/{$+16$},4/{$+24$},5/{$+32$}}{
  \node[anchor=north,font=\fC,text=inkc] at ($(B\i)+(0,-1.5pt)$) {\lab};
}
\node[anchor=north,font=\fC,text=mutedc]
     at ($(P0)!0.5!(P1)+(0,-10pt)$) {ZDP shift $\Delta$ (px)};
\end{tikzpicture}
    \setlength{\abovecaptionskip}{0.5em}
    \setlength{\belowcaptionskip}{0em}
\caption{\textbf{Moving the zero-disparity plane (ZDP) breaks existing stereo matching models.} One scene rendered at five ZDP shifts $\Delta$.\
The disparity shifts from positive only (red, in front of the ZDP) to predominantly negative (blue, behind it).
We show that SOTA FoundationStereo (the 3$^{rd}$ row) cannot express the negative regime. Our method (the 4$^{th}$ row) restores this capability.}
\label{fig:real3d_teaser}
\end{figure}
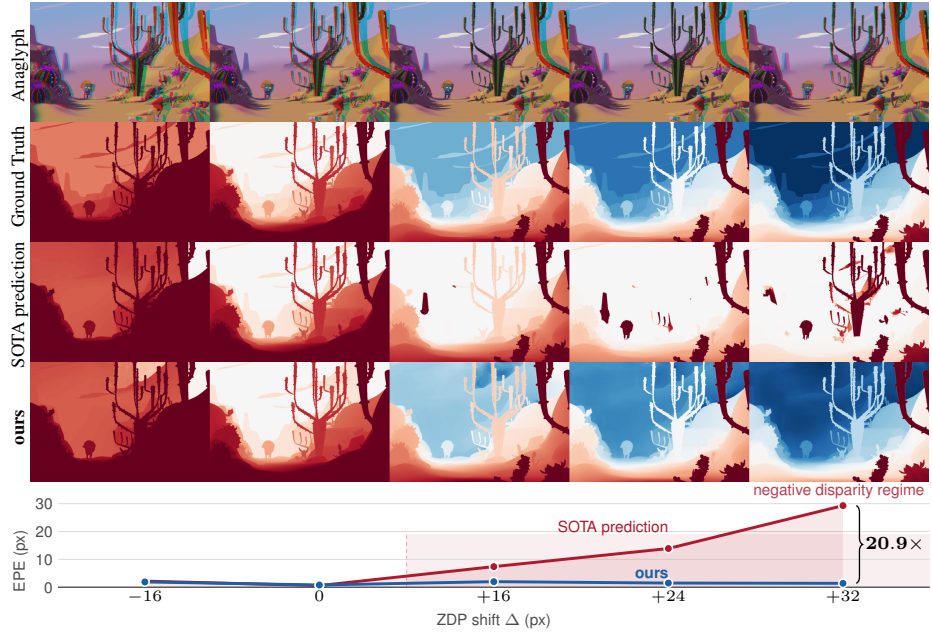

\begin{abstract}

Modern stereo matching models fail when disparity crosses zero, with end-point error (EPE) rising by 4.6–37$\times$. Yet stereoscopic content, from cinema 3D to VR, routinely contains objects behind the zero-disparity plane (ZDP), corresponding to negative disparities.
The blind spot cascades through datasets, architectures, and evaluation protocols, all of which inherit the non-negative geometry. 
Rectified parallel cameras place ZDP at infinity, so every finite depth yields $d=fB/z \ge 0$ by construction, and nothing within the standard pipeline can violate, or even measure, a negative disparity.
To measure it, we propose \textit{ZDPShift}, a benchmark of $21{,}495$ stereo pairs from seven cinematographer-authored open movies, each frame rendered at five zero-disparity-plane positions with dense signed ground truth.
Six state-of-the-art image and video stereo matching models collapse once the plane moves. On identical scene content, FoundationStereo goes from $2.24$\,px EPE to $75.33$\,px, with every backbone leaving roughly half of all pixels exceeding a three-pixel disparity error.
What is missing, however, is not the underlying matching capability.
Training on supervision synthesized from SceneFlow, which adds no new data or parameters, keeps the error flat across the signed range. Training only the decoder, with the pretrained matching features frozen, performs comparably across all six backbones, with EPE jittering within $0.2$\,px. Thus, the pretrained features already extend to the negative regime they were never trained on, and only the output convention discarded it. Meanwhile, positive-regime accuracy on KITTI, Middlebury, ETH3D, and Sintel is largely preserved.

\end{abstract}
\input{secs/paper_body_new}

\bibliography{iclr2026_conference}
\bibliographystyle{iclr2026_conference}

\clearpage
\appendix
\setcounter{table}{0}
\renewcommand\thetable{\Alph{table}}
\setcounter{figure}{0}
\renewcommand\thefigure{\Alph{figure}}
\input{secs/suppl}


\end{document}

%% file: math_commands.tex
\usepackage{amsmath,amsfonts,bm}

\def\eqref#1{equation~\ref{#1}}

\def\1{\bm{1}}

\DeclareMathAlphabet{\mathsfit}{\encodingdefault}{\sfdefault}{m}{sl}
\SetMathAlphabet{\mathsfit}{bold}{\encodingdefault}{\sfdefault}{bx}{n}



%% file: secs/teaser_grid.tex
\centering
\setlength{\tabcolsep}{0.5pt}          
\renewcommand{\arraystretch}{0}        
\newlength{\pw}\setlength{\pw}{0.193\linewidth}   


\newcommand{\rlab}[1]{\rotatebox[origin=l]{90}{\scriptsize #1}}
\newcommand{\clab}[1]{\makebox[\pw]{\scriptsize #1}}

\hspace{-3.5em}
\begin{tabular}{rc}
    \rlab{
        ~~~~~~\textbf{ours}~~~~
        ~~~~SOTA prediction
        ~~Ground Truth~~~
        ~Anaglyph
    }
     & 
     \includegraphics[width=0.85\linewidth]{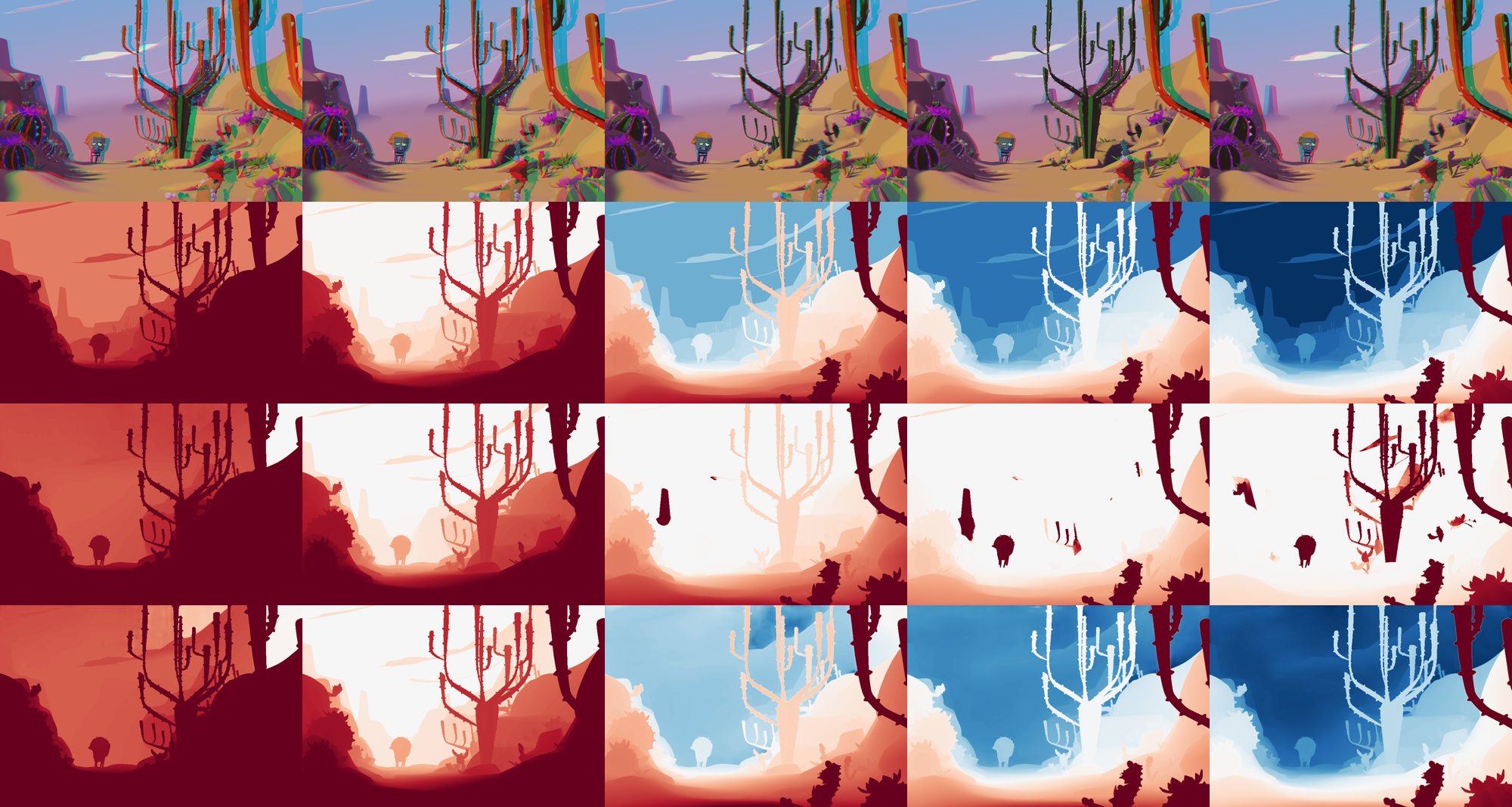}\\
\end{tabular}

%% file: secs/paper_body_new.tex

\section{Introduction}
\label{sec:introduction}

Stereoscopic content is organized around one axis: where each object sits relative to the screen. Cinema 3D, virtual reality, and head-mounted displays all manipulate the zero-disparity plane (ZDP) to place content either in front of the screen (``pop-out'') or behind it (``recede'' into the scene). Pop-out draws the eye and is used for emphasis and dramatic impact, bringing a subject out into the viewer's space, while a predominantly behind-the-screen placement opens up a deep, immersive volume that the viewer looks into, strengthening the sense of depth and presence~\citep{mendiburu2009threed}. Pushed too far in either direction, both cause visual discomfort, so artists must keep depth within a limited viewer comfort budget~\citep{shibata2011zone,lang2010nonlinear}. Signed disparity is not a corner case of stereoscopic content. It is the design language.

\begin{figure}[h]  
\centering
\renewcommand{\arraystretch}{0}
\small
\setlength{\tabcolsep}{0.05cm}
\setlength{\tabcolsep}{0pt}
\begin{tabular}{cccccc}
    \hspace{2pt}
    & \multicolumn{2}{c}{\color{teal} \textbf{FoundationStereo}}
    & \hspace{2pt}
    & \multicolumn{2}{c}{\color{violet} \textbf{StereoAnyVideo}}
    \\
    \rotatebox[origin=l]{90}{~~\textbf{Anaglyph}}
    & \includegraphics[width=.23\linewidth,trim={0 0 136cm 1cm},clip]{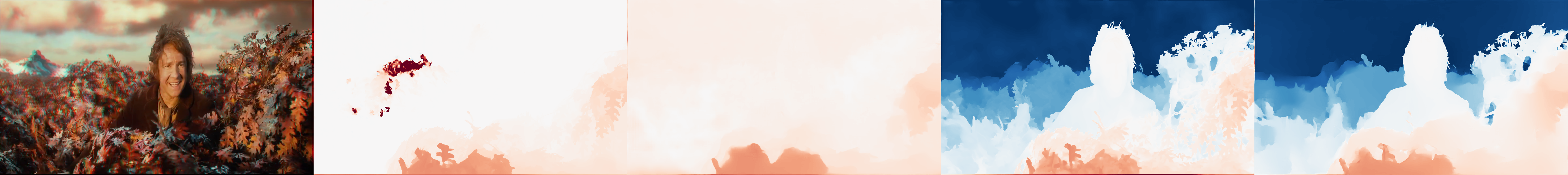}
    & \includegraphics[width=.23\linewidth,trim={0 0 136cm 1cm},clip]{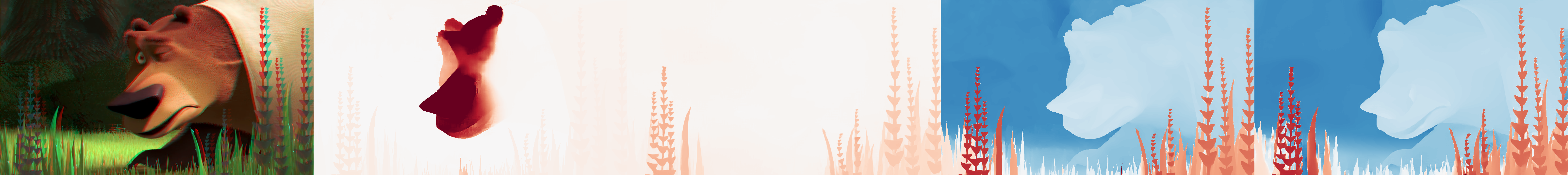}
    &
    & \includegraphics[width=.23\linewidth,trim={0 0 136cm 1cm},clip]{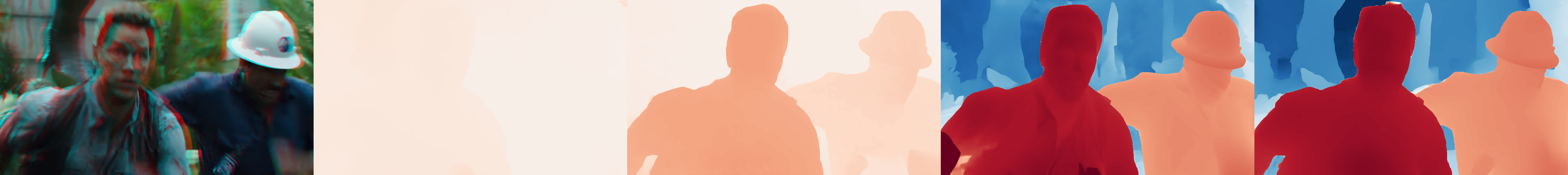}
    & \includegraphics[width=.23\linewidth,trim={0 0 136cm 1cm},clip]{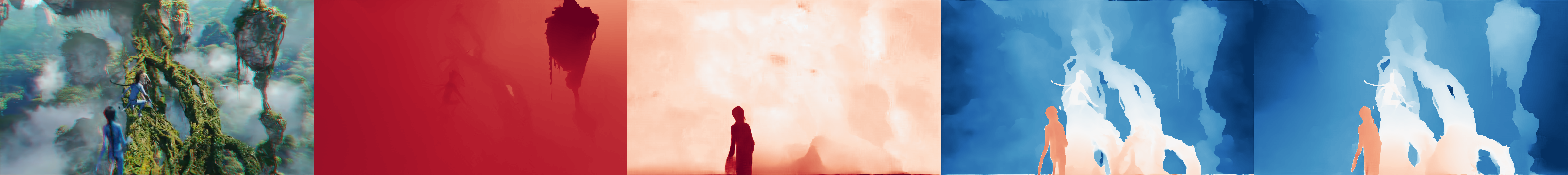}
    \\
    \rotatebox[origin=l]{90}{~\textbf{Zero-shot}}
    & \includegraphics[width=.23\linewidth,trim={34cm 0 102cm 1cm},clip]{misc/teaser/hb_p4_f8_row_rownorm.png}
    & \includegraphics[width=.23\linewidth,trim={34cm 0 102cm 1cm},clip]{misc/teaser/os_p4_f8_row_modelnorm.png}
    &
    & \includegraphics[width=.23\linewidth,trim={68cm 0 68cm 1cm},clip]{misc/teaser/jw_t2_f15_row_modelnorm.png}
    & \includegraphics[width=.23\linewidth,trim={68cm 0 68cm 1cm},clip]{misc/teaser/av_p5_f0_row.png}
    \\
    \rotatebox[origin=l]{90}{~~~~\textbf{Ours}}
    & \includegraphics[width=.23\linewidth,trim={102cm 0 34cm 1cm},clip]{misc/teaser/hb_p4_f8_row_rownorm.png}
    & \includegraphics[width=.23\linewidth,trim={102cm 0 34cm 1cm},clip]{misc/teaser/os_p4_f8_row_modelnorm.png}
    &
    & \includegraphics[width=.23\linewidth,trim={136cm 0 0 1cm},clip]{misc/teaser/jw_t2_f15_row_modelnorm.png}
    & \includegraphics[width=.23\linewidth,trim={136cm 0 0 1cm},clip]{misc/teaser/av_p5_f0_row.png}
\end{tabular}

\setlength{\abovecaptionskip}{0.7em}
\setlength{\belowcaptionskip}{-0.3em}
\caption{\textbf{Negative disparity in released 3D films.} Four stereo films (shown as anaglyph), followed by disparity predictions (red: in front of the ZDP; blue: behind). Zero-shot FoundationStereo and Stereo-Any-Video cannot represent behind-the-screen disparities. Our signed model recovers it.}
\label{fig:real3d}
\end{figure}

\begin{figure}[t]
\centering
\includegraphics[width=.85\linewidth]{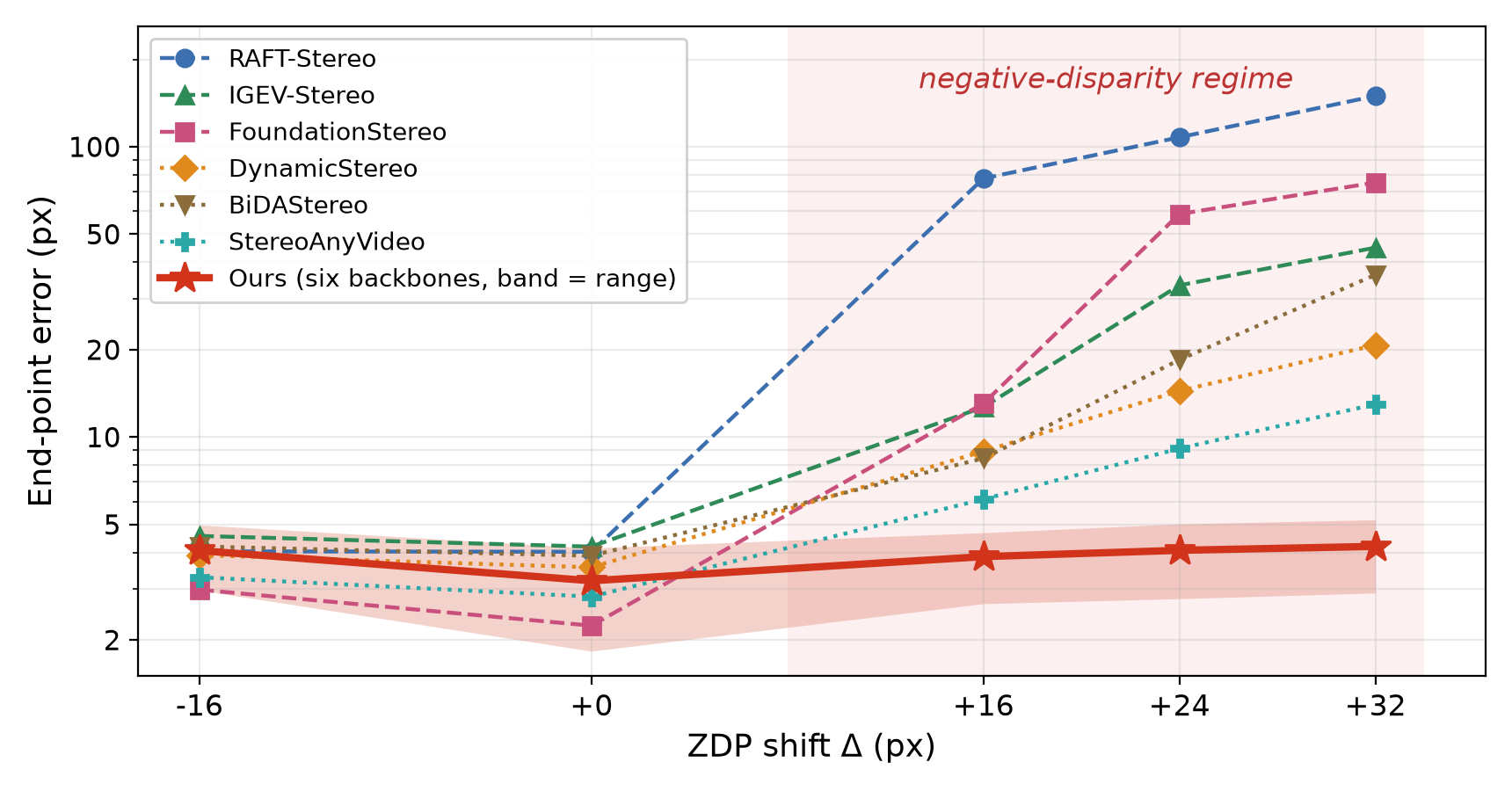}
\setlength{\abovecaptionskip}{0em}
\setlength{\belowcaptionskip}{-1em}
\caption{\textbf{End-point error (EPE) on varying the ZDP shift $\Delta$}. We evaluate on six image and video stereo matching models and our signed RAFT-Stereo (solid line). Every backbone degrades sharply once $\Delta$ grows, reaching $13$--$150$\,px EPE at $\Delta=+32$, while all six of the signed versions stay within $2.9$--$5.2$\,px across the entire signed range.}
\label{fig:final}
\end{figure}

Yet modern stereo matching cannot operate in this language. As shown in Figure~\ref{fig:real3d}, on frames from real-world stereo movies, some pixels (can be above 90\%) fall behind the screen (the ZDP), and the state-of-the-art models (FoundationStereo and Stereo-Any-Video) cannot represent this content, whereas our signed version recovers it.
Notably, ZDP is an authoring choice baked into the intrinsics as a relative principal-point offset that is never recorded in 3D movie datasets. 
Prior work on web stereo avoids this entirely, computing disparities with optical flow instead~\citep{wang2019web,xian2018monocular}. Yet, no existing benchmark contains a negative-disparity pixel, so no stereo matching model has ever been measured.
We propose the ZDPShift benchmark, a controlled multi-ZDP benchmark drawn from open-movie content.
As shown in Figure~\ref{fig:final}, we benchmark three image (RAFT-Stereo~\citep{lipson2021raft}, IGEV-Stereo~\citep{xu2023iterative}
, FoundationStereo~\citep{wen2025foundationstereo}) and three video stereo matching backbones (DynamicStereo~\citep{karaev2023dynamicstereo}, BiDAStereo~\citep{jing2024bidastereo}, StereoAnyVideo~\citep{jing2025stereo}), and find that every backbone fails the moment disparity comes with negative\footnote{We use the stereo-matching sign convention throughout: disparity $d = x_L - x_R$ is positive for scene content nearer than the zero-disparity plane and negative for content beyond it. This is opposite to the display-parallax convention common in stereoscopic production, where behind-the-screen content has positive (uncrossed) parallax and pop-out has negative (crossed) parallax~\citep{mendiburu2009threed}. Our negative-disparity regime is the behind-the-screen regime in production terms.}, with EPE increasing by $4.6$--$37\times$.
FoundationStereo, the SOTA zero-shot model at $\Delta=0$ (positive disparity only, EPE $0.94$\,px), produces an EPE of $58.1$\,px at $\Delta=+32$, which is a $62\times$ degradation on the same scene content. 
The failure is not a tail-case quirk. It occurs the instant the disparity crosses zero, and the field has no released model for negative disparities, though it has been considered essential for decades in 3D movie production.

Negative disparity is not permitted in the standard stereo matching pipeline. It is enforced through three interlocking mechanisms: \emph{(i) datasets} --- KITTI~\citep{menze2015object}, Middlebury~\citep{scharstein2014high}, ETH3D~\citep{schops2017multi}, and SceneFlow~\citep{mayer2016large} are all captured or rendered with rectified parallel rigs whose imaging geometry guarantees non-negative disparities, \emph{(ii) architectures} --- prevailing matching models encode a positive-disparity prior. This is most explicit in cost-volume methods~\citep{chang2018pyramid,guo2019group,xu2020aanet,wen2025foundationstereo}, whose correlation volume spans only a one-sided hypothesis range $[0, d_{\max}]$, making negative disparity not just unlikely but inexpressible, and \emph{(iii) evaluation} --- every standard benchmark measures only the positive regime, so the failure cannot even be observed.

We address these issues in this paper. Our contributions are:
\begin{itemize}[leftmargin=*, topsep=-3pt, itemsep=0pt]
  \item \textbf{Restoring negative disparity, cheaply.}  We remove the degradation on all six backbones. Specifically, we synthesize signed supervision by adapting SceneFlow to the full signed disparity range, and extend the one-sided hypothesis range without adding parameters. Freezing the matching features and training only the decoder can also perform reasonably well.
  \item \textbf{Measuring the negative regime.} ZDPShift is the first public stereo benchmark offering multi-ZDP renders, including $21{,}375$ pairs ($4{,}275$ open-movie frames $\times$ $5$ ZDP shifts) with signed ground truth disparity.
  \item \textbf{Signed models.}  We show that many released SOTA stereo matching models can fail with negative disparities, with EPE rising by $4.6$--$37\times$. Our resulting signed models work across the full signed range, while preserving positive-regime accuracy on standard benchmarks.
\end{itemize}

\textbf{Overview}. Section~\ref{sec:dataset} introduces the ZDPShift benchmark and its collection protocol. Section~\ref{sec:recipe} presents the architectural and data techniques that enable negative-disparity estimation. Section~\ref{sec:result} gives our results across six image and video backbones. We present discussions and conclusions in Sections~\ref{sec:discussion} and~\ref{sec:conclusion}, respectively.

\section{Related Work}
\subsection{Stereoscopic perception and ZDP placement}
Stereoscopic comfort is bounded on both sides of the zero-disparity plane~\citep{shibata2011zone,lambooij2009visual}: content too far in front of it and too far behind it is both fatiguing. A scene's depth range is fixed by its content, but where that range falls relative to the screen is not. The ZDP is the offset that maps scene depth onto the display's comfort budget, setting how much of the frame sits in front of the screen and how much behind. Stereoscopic cinematography, therefore, treats its placement as a primary control~\citep{mendiburu2009threed}, adjusted per shot alongside disparity remapping~\citep{lang2010nonlinear} or applied directly~\citep{SHAO2015125}. Automating that choice requires measuring disparity on both sides of the plane, which no released stereo matching model can do.

\subsection{Stereo matching architectures} Modern deep stereo matching models split into two dominant families. {Cost-volume aggregation} techniques (PSMNet~\citep{chang2018pyramid}, GwcNet~\citep{guo2019group}, AANet~\citep{xu2020aanet}, FoundationStereo~\citep{wen2025foundationstereo}) mostly build correlation volumes over a one-sided disparity hypothesis range $[0, d_{\max}]$ that halves memory and compute but makes negative disparity inexpressible. {Iterative recurrent} methods (RAFT-Stereo~\citep{lipson2021raft}, IGEV-Stereo~\citep{xu2023iterative}) relax the volume but initialize the disparity field at zero and train exclusively on non-negative corpora. Prior stereo work broadens competence within the non-negative regime. However, ours is orthogonal. We target the regime boundary itself, an axis no prior model has addressed.

A parallel line of work reframes two-view geometry in scene space, such as pointmap regression (DUSt3R~\citep{wang2024dust3r}), its matching-augmented successor MASt3R~\citep{leroy2024grounding}, and feed-forward multi-view transformers (VGGT~\citep{wang2025vggt}). Those methods are indifferent to the ZDP by construction. However, they operate far from the sub-pixel precision that stereoscopic use requires (EPE$> 10$px), which we quantify in the supplementary.

\subsection{Datasets} Common datasets (KITTI~\citep{geiger2012ready,menze2015object}, Middlebury~\citep{scharstein2014high}, ETH3D~\citep{schops2017multi}, SceneFlow~\citep{mayer2016large}) use rectified parallel rigs, placing the zero-disparity plane at infinity, so that negative-disparity pixels cannot occur.
As Table~\ref{tab:dataset_comparison} shows, no prior dataset renders the same scene at multiple ZDP positions.

\begin{table}[h]
    \footnotesize
    \centering
    \setlength{\abovecaptionskip}{0.5em}
    \setlength{\belowcaptionskip}{-0.5em}
    \caption{ZDPShift versus existing stereo datasets along the axes that matter for the negative-disparity regime: capture/render convention, ZDP coverage, disparity sign, and scale.}
    \label{tab:dataset_comparison}
    \setlength{\tabcolsep}{3pt}
    \begin{tabular}{lHcccHHc  HHHHlr}
    \toprule
    Dataset & Venue & Indoor & Outdoor & Dense & Accuracy & Diversity & Annotation & Baseline & FL & Range & Ave./Med. & Disp.~sign & \# images \\
    \midrule
    Sintel~\citep{Butler2012-sintel} & ECCV2012           & \checkmark & \checkmark & \checkmark & High & Medium & Synthetic & 0.1m & - & 0-972 & 66.5/25 & pos & 1,064 \\
    KITTI12~\citep{Geiger2012-kitti12} & CVPR2012      & \xmark & \checkmark & \xmark & Medium & Low & LiDAR & 0.54 m & 720 px & 4-232 & 40.1/38 & pos & 194 \\
    Middlebury~\citep{scharstein2014high-middlebury} & GCPR2014   & \checkmark & \xmark & \xmark & High & Low & LiDAR & 140-400mm & 1100px-3600px & 15-323 & 72.5/63 & pos & 15 \\
    KITTI15~\citep{Menze2015-kitti15} & CVPR2015     & \xmark & \checkmark & \xmark & Medium & Low & LiDAR & 0.54 m & 520 px & 4-230 & 35.2/33 & pos & 200 \\
    SceneFlow~\citep{mayer2016sceneflow} & CVPR2016      & \checkmark & \checkmark & \checkmark & High & High & Synthetic & 0.54 m & - & 0-10501 & 53.9/36 & pos & 35,454 \\
    ETH3D~\citep{schops2017multi-eth3d} & CVPR2017       & \checkmark & \checkmark & \xmark & High & Low & LiDAR & 59.5mm-60.4mm & 529 px-712 px & 0-62 & 13.7/10 & pos & 27 \\
    FallingThings~\citep{tremblay2018falling} & CVPRW2018 & \checkmark & \checkmark & \checkmark & High & Low & Synthetic & 6 cm & 768.2px & 7-461 & 35.2/34 & pos & 61,500 \\
    DrivingStereo~\citep{yang2019drivingstereo} & CVPR2019 & \xmark & \checkmark & \xmark & Low & High & LiDAR & 0.54 m & 1003 & 4-128 & 31.0/26 & pos & 7,751 \\
    Argoverse~\citep{wilson2023argoverse} & CVPR2019       & \xmark & \checkmark & \xmark & Low & Low & LiDAR & - & - & 0-256 & 69.1/59 & pos & 5,530 \\
    VirtualKITTI2~\citep{cabon2020virtual} & ArXiv2020  & \xmark & \checkmark & \checkmark & High & Mid & Synthetic & 0.53 m & 725px & 0-411 & 30.1/25 & pos & 21,260 \\
    InStereo2K~\citep{bao2020instereo2k} & SCIS2020      & \checkmark & \xmark & \xmark & Low & Low & Structured Light & 10 cm(5 cm) & 8 mm & 0-328 & 78.4/74 & pos & 2,010 \\
    UnrealStereo4K~\citep{Tosi2021CVPR-unrealstereo4k} & CVPR2021 & \checkmark & \checkmark & \checkmark & High & High & Synthetic & 0.2 m(0.5 m) & - & 0-1515 & 175.3/135 & pos & 8,200 \\
    Spring~\citep{mehl2023spring} & CVPR2023        & \xmark & \checkmark & \checkmark & High & Low & Synthetic & 6.5 cm & - & 0-554 & 38.1/19 & pos & 5,000 \\
    \midrule
    Ours &           & \checkmark & \checkmark & \checkmark & High & High & Synthetic & 0.1m & - & - & - & pos \& neg & 22,025 \\
    \bottomrule
    \end{tabular}
\end{table}


\section{Measuring the Signed Axis}
\label{sec:dataset}

We propose ZDPShift, a stereo matching benchmark covering the positive and negative regimes, rendered from open movies. Negative disparity arises chiefly in 3D films, so a benchmark built from film content evaluates the main application.
We render $32$ scenes drawn from seven Blender Studio open movies (Settlers, Sprite Fright, Spring, Agent 327, Charge, Project Gold, Caminandes Llamigos). Each scene is rendered at five ZDP shifts $\Delta \in \{-16, 0, +16, +24, +32\}$ pixels with a fixed baseline $B = 0.10$ m, at $1920 \times 1280$ resolution. The principal points of the left and right cameras are offset by $\mp \Delta/2$ in pixels. The physical camera positions are untouched, and the cameras remain parallel (no toe-in), preserving the rectified-pair assumption every modern model relies on. Per-pixel ground-truth disparity follows from the standard rectified-pair geometry:
\begin{equation}
d(Z, \Delta) \;=\; \frac{fB}{Z} - \Delta,
\label{eq:disp}
\end{equation}
computed analytically from the rendered depth. Each render carries the left/right RGB image and the disparity on different $\Delta$.

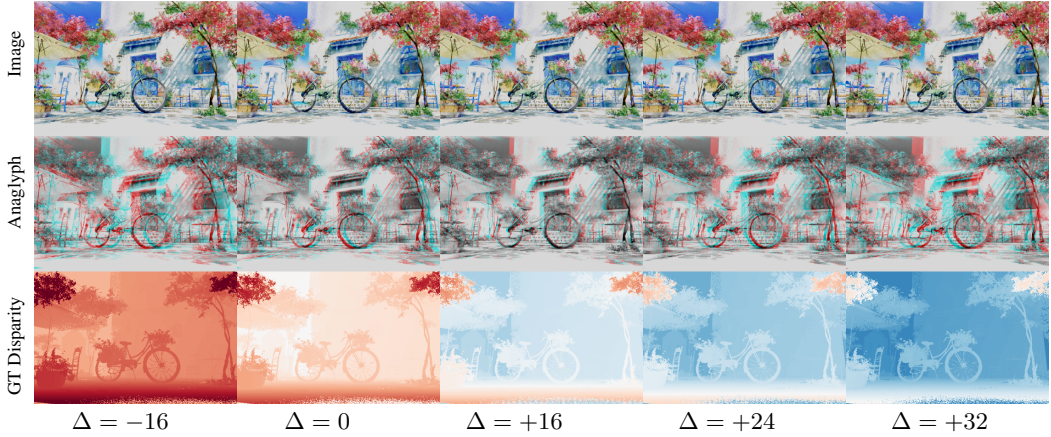
\begin{figure}[h]
    \input{secs/teaser}
    \setlength{\abovecaptionskip}{0em}
    \setlength{\belowcaptionskip}{-1em}
    \caption{
    \textbf{One scene, five zero-disparity planes}. Top$\,\to\,$bottom: reference view, red–cyan anaglyph, prediction error.
    Left$\,\to\,$right: the same scene sweeping from positive (warm color on the $3^{rd}$ row) at $\Delta=-16$ to predominantly negative (cool color on the $3^{rd}$ row) at $\Delta=+32$.
    }
  \label{fig:teaser}
\end{figure}

\subsection{Collection Protocol}
A rendered stereo pair is usable as ground truth only if every pixel has a single, sharp, photometrically consistent correspondence in the other view. Cinematic rendering is optimized for a pleasing image, not a measurable one, and breaks this requirement.
An artist's depth of field blurs out-of-focus regions, spreading a pixel's true correspondence across a circle of confusion many pixels wide. We use a narrow aperture ($f/64$) for minimal defocus effects.
Meanwhile, we bypassed the artist's compositor to disable view-dependent effects, such as lens flares.
Thus, rendered images remain sharp across the entire depth range, and the left and right views can stay photometrically consistent.
Every other factor that defines the content --- lighting, materials, scene composition, camera path, and per-shot focal length --- is inherited from the original \texttt{.blend} files unchanged. Full settings are provided in the supplementary material.

\begin{figure}[h]
\centering
\includegraphics[width=\linewidth]{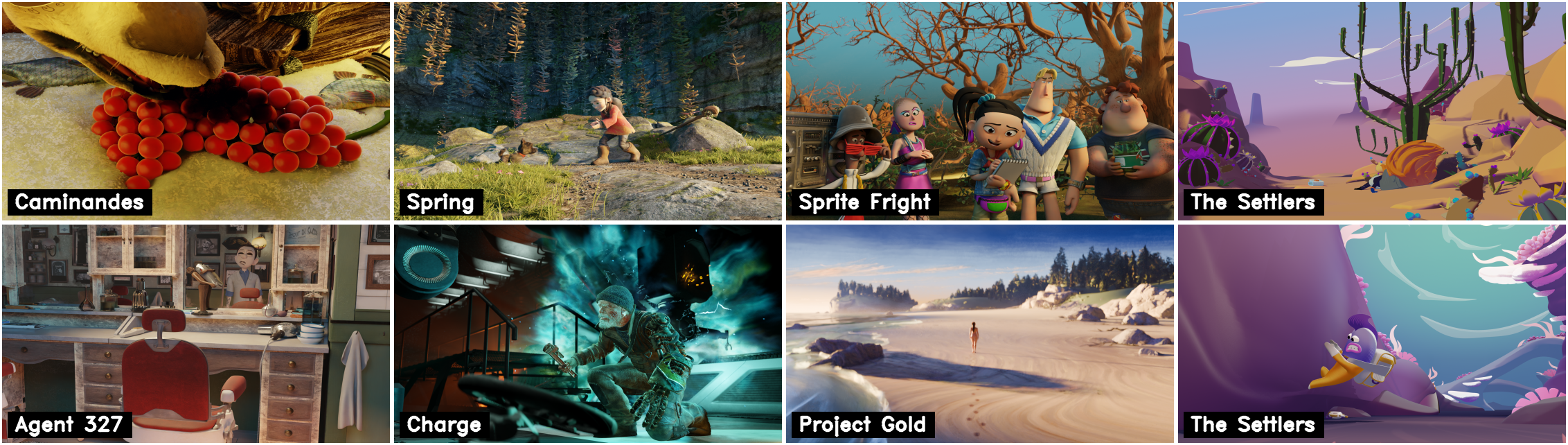}
\setlength{\abovecaptionskip}{-0.5em}
\setlength{\belowcaptionskip}{-0.5em}
\caption{\textbf{ZDPShift scene gallery}, left views. Representative frames span the seven sources, including Caminandes, Spring, Sprite Fright, Settlers, Agent 327, Charge, and Project Gold.}
\label{fig:gallery}
\end{figure}

\subsection{Dataset Statistics}

The corpus comprises $32$ scenes and $21{,}375$ stereo pairs ($4{,}275$ frames $\times$ $5$ ZDPs). All were captured at a resolution of $1920\times 1080$.
Table~\ref{tab:dataset_sources} reports how the multi-ZDP rendering populates the signed-disparity axis per source. At $\Delta\le0$ the corpus is entirely non-negative --- the regime every prior benchmark occupies --- while increasing the shift progressively converts the scene: overall $31\%$ of pixels are negative at $\Delta=+16$, rising to $40\%$ at $\Delta=+32$ and present in $78\%$ of frames, with disparities reaching $-32$\,px, all while retaining a wide positive range. The per-source breakdown shows this is a corpus-wide property rather than an artifact of a few scenes. Coverage grows monotonically with the shift for every scene, spanning content from near-screen interiors (Agent 327, essentially all-positive even at $\Delta=+32$) to deep exteriors (Project Gold, $93\%$ negative).

\begin{table}[h]
\centering
\setlength{\abovecaptionskip}{0.5em}
\setlength{\belowcaptionskip}{-0.5em}
\renewcommand{\arraystretch}{0.9}
\caption{Composition of ZDPShift (each frame rendered at five ZDP shifts) and the share of negative-disparity pixels at $\Delta>0$. Coverage varies with the content, from Agent 327's near-screen interiors to Project Gold's deep exteriors.}
\label{tab:dataset_sources}
\footnotesize
\setlength{\tabcolsep}{8pt}
\begin{tabular}{lrr rrr}
\toprule
 & & & \multicolumn{3}{c}{Neg.\ pixels (\%) at $\Delta$} \\
\cmidrule(lr){4-6}
Source & Scenes & Frames & $+16$ & $+24$ & $+32$ \\
\midrule
The Settlers       & 4 & 1{,}283 & $54$ & $59$ & $63$ \\
Sprite Fright      & 7 & 821     & $26$ & $33$ & $38$ \\
Spring             & 5 & 649     & $34$ & $39$ & $44$ \\
Agent 327          & 6 & 590     & $0$  & $0$  & $1$  \\
Charge             & 5 & 435     & $39$ & $48$ & $56$ \\
Project Gold       & 1 & 226     & $64$ & $77$ & $90$ \\
Caminandes Llamigos & 4 & 271 & $37$ & $40$ & $46$ \\
\midrule
Total              & 32 & 4{,}275 & $30$ & $35$ & $40$ \\
\bottomrule
\end{tabular}
\end{table}

\section{Enabling Negative Disparity Estimation}
\label{sec:recipe}


As discussed, for a complete system, three aspects need to be addressed: one-sided training data, one-sided hypothesis ranges, and evaluation that never crosses zero. ZDPShift addresses the third. This section addresses the other two with a pair of lightweight modifications.

\paragraph{Signed supervision from existing data.} Shifting the right image of a rectified pair by $\Delta$ pixels moves the ZDP, so relabelling the ground truth as $d' = d - \Delta$ yields a physically consistent pair whose disparities are negative wherever $d < \Delta$. We apply this to SceneFlow~\citep{mayer2016large} with $\Delta \in \{-16, 0, +16, +24, +32\}$\,px, drawn uniformly per sample, and keep half of every batch unshifted to preserve positive-disparity competence. For the video models, a single $\Delta$ is used per clip, keeping the synthesized plane constant. Two additional datasets, Dynamic Replica~\citep{karaev2023dynamicstereo} and CREStereo~\citep{li2022practical}, are included for training the video models.

\paragraph{A signed hypothesis range.} Unlike IGEV-Stereo and FoundationStereo, enabling negative disparities on RAFT-Stereo and the three video models needs no architectural change. IGEV-Stereo and FoundationStereo build a group-wise correlation (GWC) volume indexed over the non-negative range $[0, d_{\max}/4]$, which is out of bounds once the estimate goes negative. We simply extend the GWC index range to $[-d_{\text{neg}}, d_{\text{pos}}]$ and offset the soft-argmax over the signed bins. The reformulation adds no additional training parameters.

An interesting question is therefore raised: \textit{Does the signed regime demand new correspondences, or only an output convention that admits them?} To answer it, we train a variant with the matching-feature pathway \emph{frozen}, optimizing only the decoder. This will shed light on how much of the negative regime the pretrained correspondences already cover.

\paragraph{Further evaluation.} The supplementary reports evaluations of MASt3R and VGGT, manual ZDP re-positioning, training data, detailed frozen ablations, qualitative results, etc. We encourage readers to watch the accompanying videos, where the gain on real production stereo is clearest.

\section{Results}
\label{sec:result}

\subsection{Experiment Setting}
All experiments run on a single NVIDIA H200 (160\,GB) in PyTorch with FP16 mixed precision. All models are trained for $30$k iterations. 
All training uses AdamW with a OneCycle schedule (peak learning rate $2\times10^{-4}$). We use training resolutions of $320\times 480$ and $320\times 448$ for image and video models, respectively. All models infer at a fixed width of $960$. FoundationStereo's DINOv2 backbone is kept frozen, leaving $37.6$M of its $62.3$M parameters trainable. Each training completes in $40$k steps on a single H200 GPU (approx. $3$\,h for image models, $10$\,h for video models). The signed cost volume (for FoundationStereo and IGEV-Stereo) used is $d_{\text{neg}}=64$ and $d_{\text{pos}}=192$.

During inference, the disparity range is bounded to $[-128,+384]$ at a fixed width of $960$\,px (the excluded content is $1.5\%$ of valid pixels).
Temporal EPE uses one $16$-frame clip per scene.

\newcommand{\zs}[1]{\cellcolor{red!7}#1}
\begin{table}[h]
\centering
\setlength{\abovecaptionskip}{0.5em}
\setlength{\belowcaptionskip}{-0.5em}
\caption{\textbf{Quantitative comparisons on ZDPShift}.
In addition to \emph{Ours (full training)}, \emph{Ours (frozen)} keeps the pretrained matching features frozen.
Zero-shot accuracy collapses as the ZDP shifts into the negative regime, while both of our signed variants stay nearly flat with unchanged temporal consistency. Best and second-best performances are annotated in \textbf{\color{red}red} and \textbf{\color{cyan}\underline{cyan}} color, respectively. A light shade is applied to the cells where the negative-disparity regime collapsed significantly.
}
\label{tab:main}
\footnotesize
\setlength{\tabcolsep}{4pt}
\renewcommand{\arraystretch}{0.9}
\begin{tabular}{l ccccc ccccc c}
\toprule
& \multicolumn{5}{c}{EPE $\downarrow$ @ $\Delta$}
& \multicolumn{5}{c}{TEPE $\downarrow$ @ $\Delta$}
& Bad3$\downarrow$ \\
\cmidrule(lr){2-6}\cmidrule(lr){7-11}\cmidrule(lr){12-12}
Model & $-16$ & $0$ & $+16$ & $+24$ & $+32$ & $-16$ & $0$ & $+16$ & $+24$ & $+32$ & ${+}32$ \\
\midrule
\multicolumn{12}{l}{\emph{Original (zero-shot)}} \\
\rowcolor{lightgray}\multicolumn{12}{l}{Video Models} \\
\quad DynamicStereo     & $3.84$ & $3.46$ & \zs{$9.02$} & \zs{$14.61$} & \zs{$20.90$} & $1.29$ & $1.28$ & $1.59$ & $1.93$ & $2.12$ & \zs{$46.6\%$} \\
\quad BiDAStereo        & $4.12$ & $3.80$ & \zs{$8.51$} & \zs{$18.69$} & \zs{$36.62$} & $1.22$ & $1.21$ & $1.62$ & $2.12$ & $2.79$ & \zs{$47.2\%$} \\
\quad StereoAnyVideo    & $3.15$ & $2.73$ & \zs{$6.15$} & \zs{$9.17$} & \zs{$12.91$} & $1.14$ & $1.10$ & $1.18$ & $1.30$ & $1.44$ & \zs{$45.2\%$} \\
\rowcolor{lightgray}\multicolumn{12}{l}{Image Models} \\
\quad RAFT-Stereo       & $3.98$ & $3.98$ & \zs{$78.31$} & \zs{$108.41$} & \zs{$150.46$} & $1.75$ & $1.48$ & \zs{$5.98$} & \zs{$6.68$ }& \zs{$11.08$} & \zs{$51.0\%$} \\
\quad IGEV-Stereo       & $3.85$ & $3.63$ & \zs{$12.38$} & \zs{$33.11$} & \zs{$44.79$} & $1.40$ & $1.37$ & $2.84$ & \zs{$5.21$} & \zs{$6.40$} & \zs{$50.2\%$} \\
\quad FoundationStereo  & $2.91$ & $2.15$ & \zs{$13.03$} & \zs{$58.80$} & \zs{$75.53$} & $1.10$ & $1.12$ & $2.42$ & \zs{$5.61$} & \zs{$10.69$} & \zs{$47.9\%$} \\
\midrule
\multicolumn{12}{l}{\emph{Ours (frozen matching features)}} \\
\rowcolor{lightgray}\multicolumn{12}{l}{Video Models} \\
\quad DynamicStereo     & $3.98$ & $3.27$ & $3.95$ & $4.15$ & $4.35$ & $1.12$ & $1.10$ & $1.10$ & $1.10$ & $1.13$ & $18.7\%$ \\
\quad BiDAStereo        & $4.68$ & $4.02$ & $4.30$ & $4.67$ & $4.75$ & $1.10$ & $1.10$ & $1.09$ & $1.10$ & $1.09$ & $18.1\%$ \\
\quad StereoAnyVideo     & $3.67$ & $2.74$ & $3.56$ & $3.77$ & $4.06$ & $\color{cyan}\mathbf{\underline{1.03}}$ & $\color{cyan}\mathbf{\underline{1.03}}$ & $\color{cyan}\mathbf{\underline{1.01}}$ & $\color{cyan}\mathbf{\underline{1.02}}$ & $\color{cyan}\mathbf{\underline{1.02}}$ & $18.4\%$ \\
\rowcolor{lightgray}\multicolumn{12}{l}{Image Models} \\
\quad RAFT-Stereo       & $5.01$ & $3.96$ & $4.87$ & $4.81$ & $4.78$ & $1.87$ & $1.76$ & $1.71$ & $1.76$ & $1.67$ & $22.3\%$ \\
\quad IGEV-Stereo       & $4.36$ & $3.53$ & $3.98$ & $3.85$ & $3.85$ & $1.80$ & $1.72$ & $1.55$ & $1.63$ & $1.47$ & $19.5\%$ \\
\quad FoundationStereo  & {$\color{red}\mathbf{2.82}$} & $\color{cyan}\mathbf{\underline{1.82}}$ & $\color{cyan}\mathbf{\underline{2.71}}$ & $\color{cyan}\mathbf{\underline{2.78}}$ & $\color{cyan}\mathbf{\underline{2.94}}$ & $1.13$ & $1.05$ & $1.08$ & $1.09$ & $1.08$ & $\color{cyan}\mathbf{\underline{16.0\%}}$ \\
\midrule
\multicolumn{12}{l}{\emph{Ours (full training)}} \\
\rowcolor{lightgray}\multicolumn{12}{l}{Video Models} \\
\quad DynamicStereo     & $4.16$ & $3.39$ & $3.95$ & $4.30$ & $4.75$ & $1.12$ & $1.11$ & $1.10$ & $1.10$ & $1.12$ & $18.2\%$ \\
\quad BiDAStereo        & $4.54$ & $3.85$ & $4.16$ & $4.41$ & $4.56$ & $1.11$ & $1.09$ & $1.10$ & $1.11$ & $1.11$ & $18.0\%$ \\
\quad StereoAnyVideo    & $3.69$ & $2.75$ & $3.39$ & $3.66$ & $3.90$ & $\color{red}\mathbf{{1.01}}$ & $\color{red}\mathbf{1.00}$ & $\color{red}\mathbf{{0.99}}$ & $\color{red}\mathbf{0.99}$ & $\color{red}\mathbf{{1.00}}$ & $17.1\%$ \\
\rowcolor{lightgray}\multicolumn{12}{l}{Image Models} \\
\quad RAFT-Stereo       & $4.90$ & $4.06$ & $4.60$ & $4.55$ & $4.57$ & $1.91$ & $1.85$ & $1.83$ & $1.82$ & $1.77$ & $20.9\%$ \\
\quad IGEV-Stereo       & $4.12$ & $3.07$ & $3.69$ & $3.69$ & $3.73$  & $1.91$ & $1.71$ & $1.56$ & $1.60$  & $1.48$ & $19.0\%$ \\
\quad FoundationStereo  & $\color{cyan}\mathbf{\underline{2.91}}$ & {$\color{red}\mathbf{1.77}$} & {$\color{red}\mathbf{2.61}$} & {$\color{red}\mathbf{2.72}$} & {$\color{red}\mathbf{2.84}$} & $1.15$ & $1.05$ & $1.06$ & $1.09$ & $1.03$ & $\color{red}\mathbf{15.7\%}$ \\
\bottomrule
\end{tabular}
\end{table}

\subsection{Quantitative Results}

\paragraph{Every backbone degrades steeply beyond the zero-disparity plane.} Wherever the plane leaves every disparity positive ($\Delta \le 0$), every backbone performs well. At $\Delta=+32$, the same models on the same scenes degrade by around $4$--$37\times$, with roughly half of all pixels exceeding a three-pixel disparity error. Notably, moving towards the positive side ($\Delta=-16$) does not break the performance. However, the error increases monotonically as $\Delta$ increases for all six backbones.
The negative regime is outside what they can express.

\begin{wrapfigure}{r}{0.49\textwidth}
  \vspace{-1.5\intextsep}
  \begin{center}
    \includegraphics[width=0.48\textwidth]{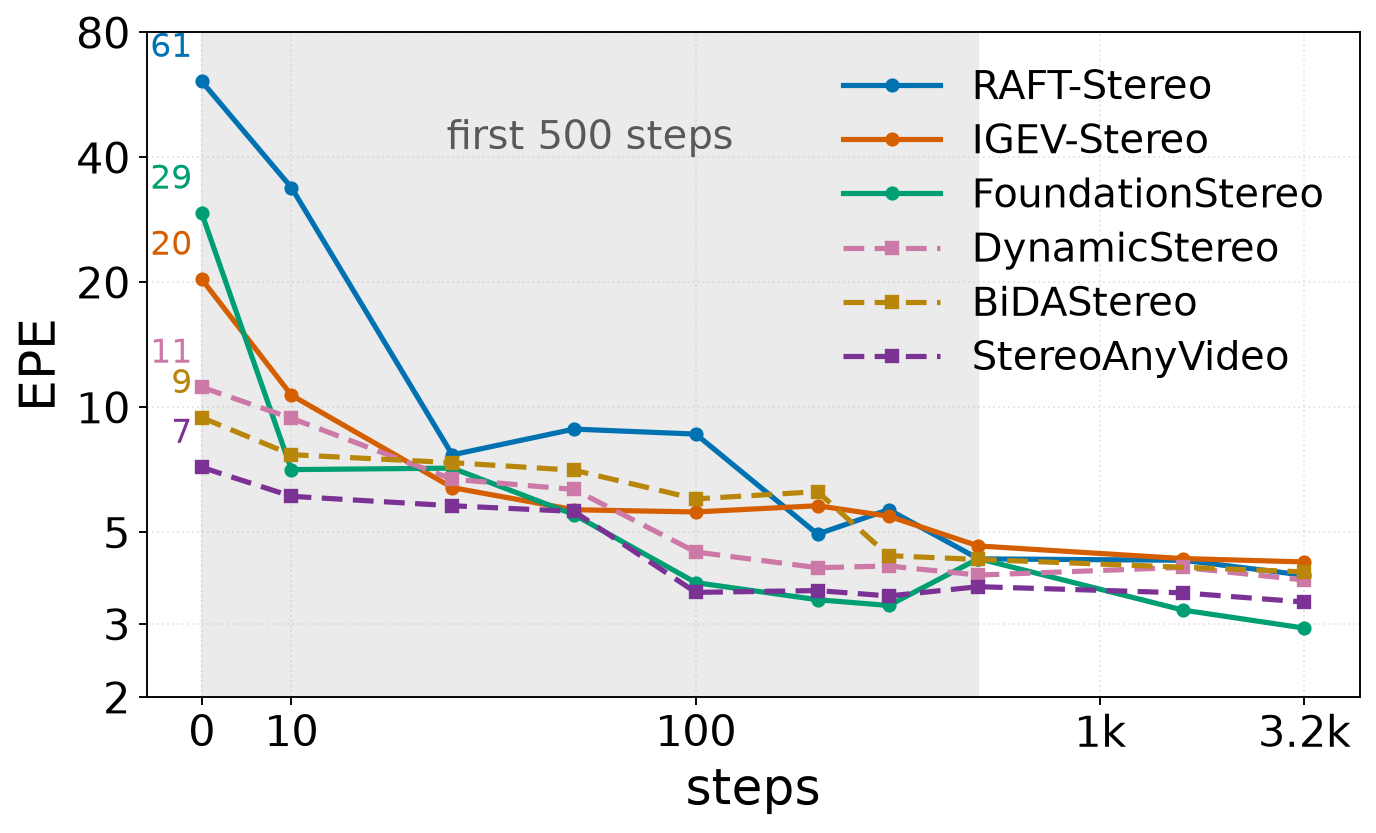}
  \end{center}
  \setlength{\abovecaptionskip}{0em}
  \setlength{\belowcaptionskip}{-2em}
  \caption{\textbf{The negative-disparity capability surfaces almost immediately.} Both axes are logarithmic, and step~0 is the pretrained model.}
  \label{fig:converge}
\end{wrapfigure}

\paragraph{The features already support negative disparity.}  Although our method flattens the signed axis for every backbone, we observe that the frozen variant matches full training on all six backbones, within $0.2$\,px throughout, and temporal consistency is equally unaffected. This indicates that the matching features can already deal with negative disparity.
We use a fixed 800-pair sample of ZDPShift as the evaluation protocol. 
As shown in Figure~\ref{fig:converge}, models with a frozen matching feature extractor can achieve comparable performance within $500$ steps, indicating that the features computed by the pretrained models can match content behind the ZDP. 
Thus, the representation from pretrained models is already sign-agnostic, which transfers to disparities they never saw.

\paragraph{The positive regime is preserved.} We evaluate on six standard benchmarks containing no negative disparity: KITTI-2015~\citep{menze2015object}, Middlebury~\citep{scharstein2014high}, ETH3D~\citep{schops2017multi} and Sintel~\citep{Butler2012-sintel} for the image models, and Sintel and Virtual KITTI\,2~\citep{cabon2020virtual} for the video models. In general, the cost of gaining the signed regime is small and mostly comparable. Set against the gain in the negative regime, the trade is favourable for every backbone.
\begin{table}[h]
\centering
\setlength{\abovecaptionskip}{0.2em}
\setlength{\belowcaptionskip}{-0.3em}
\caption{\textbf{Positive-regime generalization across standard benchmarks.} We compare the original model (Src) against ours.  Ours retains competitive positive-regime accuracy and temporal consistency, with negative capability enabled.}
\label{tab:external}
\footnotesize
\begin{tabular}{lcccc cccc}
\toprule
 & \multicolumn{4}{c}{Sintel} & \multicolumn{4}{c}{Virtual KITTI\,2} \\
\cmidrule(lr){2-5}\cmidrule(lr){6-9}

\cellcolor{lightgray}{Video Models} & \multicolumn{2}{c}{EPE$\downarrow$} & \multicolumn{2}{c}{TEPE$\downarrow$}
& \multicolumn{2}{c}{EPE$\downarrow$} & \multicolumn{2}{c}{TEPE$\downarrow$} \\
\cmidrule(lr){2-3}\cmidrule(lr){4-5}
\cmidrule(lr){6-7}\cmidrule(lr){8-9}
& Src & Ours & Src & Ours & Src & Ours & Src & Ours \\
\midrule
DynamicStereo  & $1.95$ & $\mathbf{1.69}$ & $1.02$ & $\mathbf{0.98}$ & $1.71$ & $\mathbf{1.61}$ & $1.54$ & $\mathbf{1.52}$ \\
BiDAStereo     & $\mathbf{1.52}$ & $1.72$ & $\mathbf{0.92}$ & $0.98$ & $1.68$ & $\mathbf{1.64}$ & $\mathbf{1.49}$ & $1.52$ \\ 
StereoAnyVideo & $\mathbf{1.47}$ & $1.53$ & $\mathbf{0.86}$ & $\mathbf{0.86}$ & $\mathbf{1.12}$ & $1.29$ & $\mathbf{1.11}$ & $1.22$ \\
\bottomrule
\end{tabular}
\hfill
\begin{tabular}{lcc cc cc cc}
\toprule
& \multicolumn{2}{c}{KITTI2015}
& \multicolumn{2}{c}{Middlebury}
& \multicolumn{2}{c}{ETH3D}
& \multicolumn{2}{c}{Sintel} \\
\cmidrule(lr){2-3}\cmidrule(lr){4-5}\cmidrule(lr){6-7}\cmidrule(lr){8-9}
\cellcolor{lightgray}{Image Models} 
& \multicolumn{2}{c}{EPE$\downarrow$}
& \multicolumn{2}{c}{EPE$\downarrow$}
& \multicolumn{2}{c}{EPE$\downarrow$}
& \multicolumn{2}{c}{EPE$\downarrow$}\\
\cmidrule(lr){2-3}\cmidrule(lr){4-5}\cmidrule(lr){6-7}\cmidrule(lr){8-9}
& Src & Ours & Src & Ours & Src & Ours & Src & Ours \\
\midrule
RAFT-Stereo      & $\mathbf{1.32}$ & $1.76$ & $\mathbf{2.09}$ & $2.33$ & $0.42$ & $\mathbf{0.39}$ & $2.01$ & $\mathbf{1.94}$ \\
IGEV-Stereo      & $\mathbf{1.38}$ & $\mathbf{1.38}$ & $2.01$ & $\mathbf{1.58}$ & $0.34$ & $\mathbf{0.27}$ & $1.92$ & $\mathbf{1.59}$ \\
FoundationStereo & $\mathbf{0.97}$ & $1.07$ & $\mathbf{0.92}$ & $0.98$ & $\mathbf{0.17}$ & $\mathbf{0.17}$ & $1.22$ & $\mathbf{1.06}$ \\
\bottomrule
\end{tabular}
\end{table}

\paragraph{Are the negative disparity offsets memorized?}
Instead of training and inference on a fixed set of $\Delta$, we repeat the experiment with $\Delta{=}{+}32$ withheld, training on the remaining four for $20$k steps and keeping ${+}32$ unseen.
As shown in Table~\ref{tab:heldout32}, models achieve comparable performance on unseen shifts, not simply memorizing training offsets.
\begin{table}[h]
\centering
\setlength{\abovecaptionskip}{0.5em}
\setlength{\belowcaptionskip}{-0.3em}
\caption{\textbf{Generalization to an unseen shift}, with $\Delta{=}{+}32$ withheld from training. EPE$_{-16\ldots{+}24}$ denotes the mean EPE of the predictions for $\Delta\in\{-16,0,8, 16,24\}$.}
\label{tab:heldout32}
\small
\setlength{\tabcolsep}{3pt}
\begin{tabular}{l rrr rrr}
\toprule
& \multicolumn{3}{c}{Trained on all $\Delta$} & \multicolumn{3}{c}{$\Delta = {+}32$ withheld} \\
\cmidrule(lr){2-4}\cmidrule(lr){5-7}
Model & EPE$_{-16\ldots{+}24}$ & EPE$_{{+}32}$ & TEPE$_{{+}32}$ & EPE$_{-16\ldots{+}24}$ & EPE$_{{+}32}$ & TEPE$_{{+}32}$ \\
\midrule
RAFT-Stereo       & 4.53 & 4.57 & 1.93 & 4.21 & 4.31 & 1.60 \\
IGEV-Stereo       & 3.64 & 3.73 & 1.48 & 3.89 & 3.96 & 1.41 \\
FoundationStereo  & 2.50 & 2.84 & 1.03 & 2.64 & 3.05 & 1.04 \\
DynamicStereo & 4.28 & 5.02 & 1.14 & 4.02 & 4.95 & 1.11 \\
BiDAStereo & 3.92 & 4.38 & 1.15 & 3.90 & 4.36 & 1.14 \\
StereoAnyVideo & 3.28 & 3.89 & 1.03 & 3.22 & 3.81 & 0.99 \\
\bottomrule
\end{tabular}
\end{table}

\section{Discussion}
\label{sec:discussion}

Measuring $81$ clips ($16$ frames each) from ten released stereoscopic films with our signed models, $89\%$ of frames contain negative disparity, and $67\%$ contain more than half of the pixels in the negative regime (see supplementary for per-film distribution). 
The stereo matching field, however, grew up on a more controlled geometry. The rectified parallel rig fixes the ZDP at infinity, producing only positive disparity by construction.
Thus, the convention hardened into datasets, hypothesis spaces, and evaluation protocols, where it became invisible as the standard stereo matching pipeline.

Freezing the matching features comes within a $0.2$\,px difference of full tuning everywhere. Note that those matching features are never updated, yet they support correspondences of a sign they never trained on. Thus, what these matching features encode is the correspondence, and the sign convention lives in the output layer, not in the representation.

\paragraph{ZDP for visual comfort.} The degree of visual discomfort can be predicted from a few factors, such as spatial frequency, disparity response, and visual attention~\citep{SHAO2015125}. 
During production, artists tend to place ZDPs to keep the whole video stereoscopic and within the viewer's comfort budget~\citep{shibata2011zone,lambooij2009visual}; because scene depth evolves within and across shots, that placement cannot stay fixed.
Comfort budget bounds the parallax angle to about $1^\circ$ either side of the screen~\citep{lambooij2009visual}.~\footnote{At the standard viewing distance of three screen heights on a $16{:}9$ display, $1^\circ$ subtends $3 \cdot \tan(1^\circ)\,W/1.78 \approx 0.029\,W$, i.e.\ $2.9\%$ of screen width.}
The budget is itself defined on signed disparity: how far content sits in front of the screen and how far behind.
Thus, with models that never output negative disparity, stereo comfort on real films cannot be measured. A signed stereo matching model makes it computable, turning ZDP placement from a judgment made by hand into a quantity that can be measured, scored, and optimized over a sequence.

\definecolor{zdpInk}{HTML}{1A1A1A}      
\definecolor{zdpFront}{HTML}{C1553B}    
\definecolor{zdpBehind}{HTML}{2C6E9B}   
\definecolor{zdpBand}{HTML}{EDE7DE}     
\definecolor{zdpGrid}{HTML}{D8D8D4}

\begin{wrapfigure}{r}{0.44\textwidth}
\vspace{-1.5\intextsep}
\centering
\begin{tikzpicture}
\begin{axis}[
    width=1.02\linewidth, height=0.80\linewidth,
    scale only axis=false,
    xlabel={zero-disparity plane $t$},
    ylabel={MEt3R $\downarrow$},
    xmin=-0.07, xmax=1.07, ymin=0.0725, ymax=0.0975,
    xtick={0,0.25,0.5,0.75,1},
    xticklabels={$0$,$0.25$,$0.5$,$0.75$,$1$},
    ytick={0.075,0.080,0.085,0.090,0.095},
    scaled y ticks=false,
    yticklabel style={/pgf/number format/fixed, /pgf/number format/precision=3},
    axis line style={zdpInk!55, line width=0.4pt},
    axis lines=left, tick align=outside,
    tick style={zdpInk!45, line width=0.4pt},
    ymajorgrids, grid style={zdpGrid, line width=0.3pt},
    xlabel near ticks, ylabel near ticks,
    label style={font=\scriptsize\itshape, color=zdpInk},
    tick label style={font=\tiny, color=zdpInk!75},
    clip=false,
]
\addplot[draw=none, fill=zdpBand, fill opacity=0.75, forget plot]
    coordinates {(0.5,0.0725) (1.07,0.0725) (1.07,0.0975) (0.5,0.0975)} \closedcycle;
\draw[zdpInk!25, dashed, line width=0.35pt] (axis cs:0.5,0.0725) -- (axis cs:0.5,0.0975);

\addplot[draw=none, fill=zdpBehind, fill opacity=0.16, forget plot]
    coordinates {(0,0.0902) (0.25,0.0899) (0.5,0.0918) (0.75,0.0938) (1,0.0975)
                 (1,0.0825) (0.75,0.0774) (0.5,0.0746) (0.25,0.0725) (0,0.0725)}
    \closedcycle;

\addplot[color=zdpBehind, line width=1.1pt, mark=*, mark size=1.9pt,
         mark options={fill=white, draw=zdpBehind, line width=0.8pt}]
    coordinates {(0.00,0.0808) (0.25,0.0808) (0.50,0.0832) (0.75,0.0856) (1.00,0.0909)};

\node[circle, fill=zdpFront, inner sep=1.4pt] at (axis cs:0,0.0808) {};
\node[circle, fill=zdpBehind, inner sep=1.4pt] at (axis cs:1,0.0909) {};
\node[anchor=south west, font=\tiny\itshape, color=zdpFront!85!black]
    at (axis cs:-0.05,0.0814) {in front};
\node[anchor=south east, font=\tiny\itshape, color=zdpBehind!85!black]
    at (axis cs:1.03,0.0916) {behind};

\node[anchor=north east, font=\tiny, color=zdpInk!60]
    at (axis cs:1.04,0.0805) {$+12.5\%$};
\end{axis}
\end{tikzpicture}
\setlength{\abovecaptionskip}{-1em}
\setlength{\belowcaptionskip}{0em}
\caption{\textbf{Generation degrades in negative regimes.} The band is one standard error.}
\label{fig:generation}
\vspace{-\intextsep}
\end{wrapfigure}
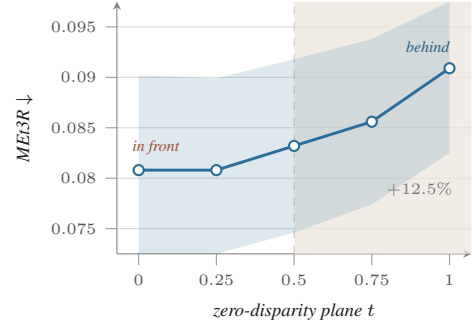
\paragraph{Stereo generation shares the blind spot.} We evaluate the ZDP tolerance on stereo generation models, where the same source frames are synthesized entirely with positive disparities ($t=0$) or entirely negative ($t=1$) with StereoCrafter~\citep{zhao2024stereocrafter}.
As shown in Figure~\ref{fig:generation}, over ten clips synthesized at five plane positions ($50$ generations), MEt3R~\citep{asim2025met3r} degrades from $0.081$ at $t{=}0$ to $0.091$ at $t{=}1$ ($+12.5\%$). $9/10$ clips are individually worse at $t{=}1$.
Details can be found in the supplementary material.
As behind-screen content is essential to viewing comfort, improving synthesis quality in the negative regime is a necessary next step for stereo video generation.


\paragraph{Limitations.}

First, a signed output space admits a new failure mode. On ambiguous, low-texture content (\eg fog or haze), a stereo matching model can occasionally flip the sign of a region, making the disparity of the low-texture area jitter. 
This accounts for most of the small residual cost we observe on real stereo films.
Second, signed ground truth exists only for rendered content. Yet, the performance on real delivered stereo (Figure~\ref{fig:real3d}) can only be validated qualitatively, since released films ship no disparity ground truth. Future work on real-world captured stereo content with negative disparities would accelerate both the benchmarking and its adoption in stereoscopic production tooling.

\section{Conclusion}
\label{sec:conclusion}

Negative disparity is not an edge case, but the regular operating regime of stereoscopic displays and movies. Surprisingly, no prior released stereo matching model handles it. 
Not because signed matching is hard, but because two one-sided conventions make it unobservable: 1) training data where $d>0$ holds by construction, and hypothesis ranges where $d<0$ are inexpressible. Both lift cheaply. Interestingly, we show that freezing the matching features still recovers most of the gain, showing that the representation could already match across the full signed-disparity range and only lacked a way to express the sign. ZDPShift makes the regime measurable, and we hope it makes signed disparity a standard axis of stereo evaluation.

%% file: secs/teaser.tex
\centering
\footnotesize
\begin{subfigure}[T]{0.03\textwidth}
    \begin{tabular}{c}
        \rotatebox{90}{\parbox{1cm}{\scriptsize Image}} \\
        \rotatebox{90}{\parbox{1.9cm}{\scriptsize Anaglyph}} \\
        \rotatebox{90}{\parbox{2.1cm}{\scriptsize GT Disparity}} \\
    \end{tabular}
\end{subfigure}
\hfill
\begin{subfigure}[T]{0.96\textwidth}
    \centering
     \includegraphics[width=1\linewidth,trim={0 5.9cm 0 0},clip]{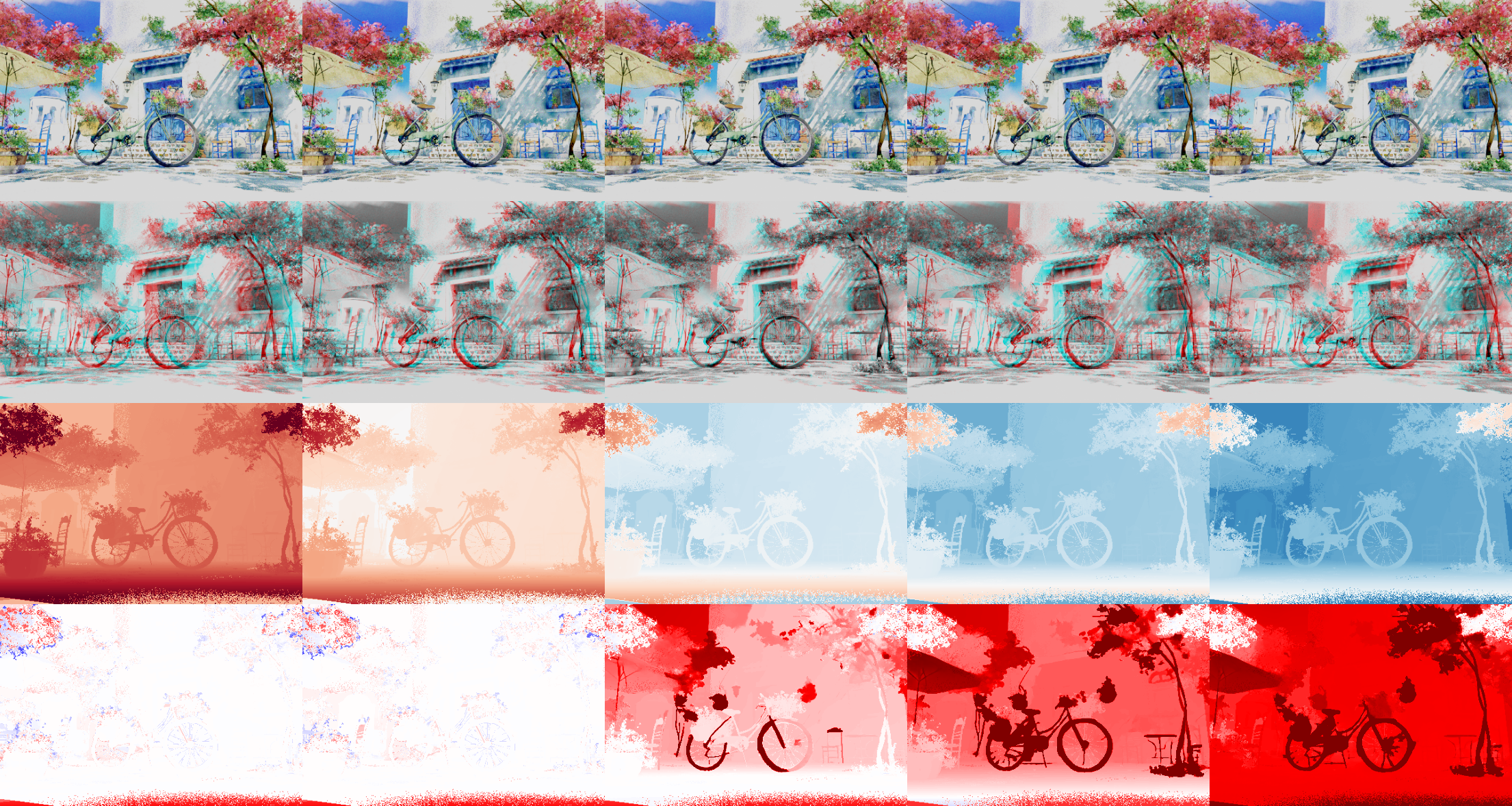}
     \begin{tabular*}{.92\linewidth}{@{\extracolsep{\fill}}ccccc}
         $\Delta=-16$ & $\Delta=0$ & $\Delta=+16$ & $\Delta=+24$ & $\Delta=+32$ \\
     \end{tabular*}
    \label{fig:sub-b}
\end{subfigure}

%% file: secs/suppl.tex


\section{Dataset generation details}
\label{sec:supp_protocol}

\paragraph{ZDP-shift implementation.} The $\mp\Delta/2$ principal-point offset of the main paper is realised in Blender as a symmetric horizontal sensor shift (off-axis projection), $\mathrm{shift}_{x,L} = +\Delta/(2W)$ and $\mathrm{shift}_{x,R} = -\Delta/(2W)$ in normalised sensor units ($W$ the image width in pixels), applied to two parallel camera copies of the artist's camera. The shift places the zero-disparity plane at depth
\begin{equation}
Z_{\mathrm{ZDP}}(\Delta) \;=\; \frac{fB}{\Delta},
\end{equation}
\textit{e.g.}\ $8.44$\,m for $f=2{,}700$\,px at $\Delta=+32$; $\Delta=0$ recovers the ZDP-at-infinity setting of existing benchmarks, and $\Delta<0$ places it virtually behind the viewer, guaranteeing a positive margin of at least $|\Delta|$\,px. Focal lengths are shot-native and span $f \in [1{,}234,\ 12{,}000]$\,px across the corpus.

\paragraph{Render settings.} All source productions are Blender Studio open movies (CC-BY). All frames were rendered with Blender~5.2.0~LTS on an Nvidia RTX 3090 GPU. Cycles shots use $32$ samples with adaptive sampling and OptiX (fallback OpenImageDenoise) denoising guided by albedo and normal passes, and persistent scene data across frames; EEVEE-authored shots render in EEVEE-Next. The forced deep aperture of the main protocol is implemented as $f/64$ with the artist's focus distance retained.
The empirical sampler-noise floor is $\approx 0.005/255$ per pixel. 

Disparity is stored per pixel as signed \texttt{float32}. No pixels are masked, every pixel carries a finite label.
The frames containing pixels at infinite depth (sky, world background) have been excluded during our post-rendering check.

\section{Negative Disparities In Production Stereo Movies}

\begin{figure}[h]
\centering
\includegraphics[width=0.92\linewidth]{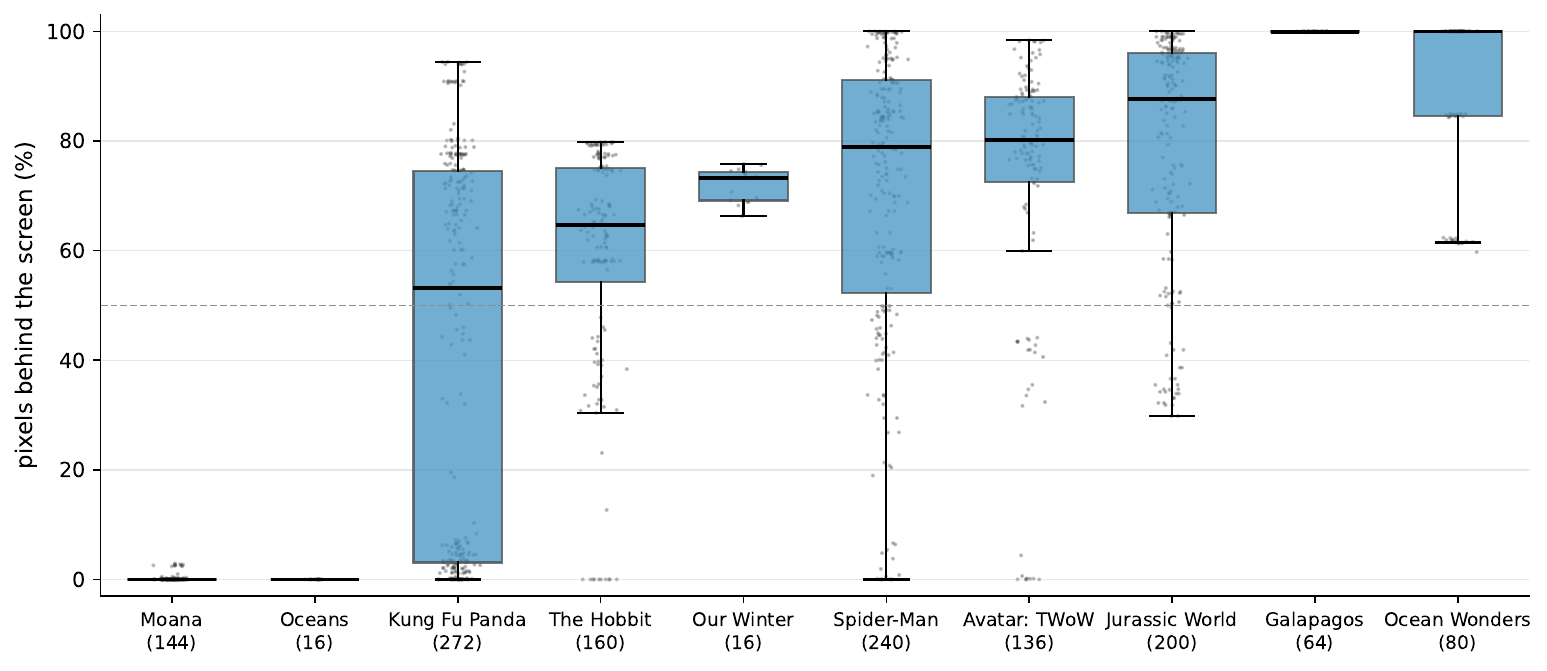}
\caption{\textbf{Per-frame share of pixels behind the screen} in ten released stereoscopic films, ordered by median. Each dot is one frame, and boxes give the per-film quartiles; the dashed line marks half the frame. Films are authored across the whole range, and most of them place the majority of the frame in the regime that no released matcher can estimate.}
\label{fig:film_neg}
\end{figure}

Released films ship no disparity ground truth, so we measure the signed distribution with our corrected FoundationStereo. 
Surprisingly, we found that negative disparity is the common case rather than the exception. It appears in $91.4\%$ of clips and covers more than half the frame in $66.7\%$; the median clip is $73.3\%$ behind the screen. Table~\ref{tab:film_neg} and Figure~\ref{fig:film_neg} give the per-film breakdown, which is wide: two documentaries are almost entirely behind the screen, while \emph{Moana} and \emph{Oceans} are authored without negative disparity at all.

We take $81$ clips from ten stereoscopic films, each a $16$-frame window, and count a pixel as behind the screen when its disparity falls below $-0.5$\,px, the threshold used for the benchmark's negative-pixel share. Frames within a window are consecutive frames of one shot and vary little. We report per-clip statistics.

\begin{table}[h]
\centering
\caption{\textbf{Negative-disparity share per film}, one value per clip (its median frame). Percentages are of image pixels behind the screen plane, from our corrected FoundationStereo.}
\label{tab:film_neg}
\small
\setlength{\tabcolsep}{8pt}
\begin{tabular}{l r rrr}
\toprule
Film & Clips & Median & $p_{25}$ & $p_{75}$ \\
\midrule
Ocean Wonders                & $5$  & $100.0$ & $84.7$ & $100.0$ \\
Galapagos                    & $4$  & $100.0$ & $99.9$ & $100.0$ \\
Jurassic World               & $12$ & $86.0$  & $63.2$ & $95.6$ \\
Spider-Man                   & $15$ & $82.5$  & $66.4$ & $89.6$ \\
Avatar: The Way of Water     & $7$  & $78.6$  & $60.0$ & $86.8$ \\
Our Winter                   & $1$  & $73.3$  & $73.3$ & $73.3$ \\
The Hobbit                   & $10$ & $65.4$  & $46.9$ & $76.7$ \\
Kung Fu Panda                & $17$ & $55.6$  & $4.0$  & $72.8$ \\
Moana                        & $9$  & $0.0$   & $0.0$  & $0.0$ \\
Oceans                       & $1$  & $0.0$   & $0.0$  & $0.0$ \\
\midrule
All                          & $81$ & $73.3$  & $21.1$ & $88.9$ \\
\bottomrule
\end{tabular}
\end{table}

\section{The blind spot on the generation side}
\label{sec:supp_generation}

By default, stereo generation places the zero-disparity plane at infinity, as stereo estimation assumes. StereoCrafter~\citep{zhao2024stereocrafter} normalizes predicted depth across a clip and maps it to a symmetric disparity range, which puts the plane in the middle of the depth range and sends roughly half of every frame behind the screen. The plane can be moved: with $t$ selecting where it falls in the normalized depth range, $t=0$ places the whole scene in front of the screen (the default setting) and $t=1$ places it all behind. The depth budget is unchanged throughout; only the sign distribution of the generated disparity moves.

We sweep $t$ over ten clips and score each synthesized pair with MEt3R~\citep{asim2025met3r}, which measures the geometric consistency of two views without ground truth. Consistency degrades monotonically as content moves behind the screen, from $0.0808$ at $t=0$ to $0.0909$ at $t=1$, a $12.5\%$ increase in the metric; in a paired comparison on the same source frames, nine of the ten clips are worse at $t=1$ than at $t=0$.

\begin{table}[h]
\centering
\caption{\textbf{Generation quality against the position of the zero-disparity plane.} MEt3R (lower is better) on ten clips synthesized by StereoCrafter, sweeping the plane from entirely in front of the screen ($t=0$) to entirely behind it ($t=1$). The source frames and the depth budget are identical across columns.}
\label{tab:generation}
\small
\setlength{\tabcolsep}{10pt}
\begin{tabular}{lccccc}
\toprule
 & \multicolumn{5}{c}{ZDP placement $t$} \\
\cmidrule(lr){2-6}
 & $0.00$ & $0.25$ & $0.50$ & $0.75$ & $1.00$ \\
 & \emph{all in front} & & \emph{half-half} & & \emph{all behind} \\
\midrule
MEt3R $\downarrow$ & $0.0808$ & $0.0808$ & $0.0832$ & $0.0855$ & $0.0909$ \\
\bottomrule
\end{tabular}
\end{table}

\begin{figure}
\centering
\setlength{\tabcolsep}{1pt}
\begin{tabular}{ccc}
\hspace{2em}
\makecell[cc]{Input Frames\\(shown as anaglyph)}\hspace{2em}
&
\raisebox{-.5\height}{
\includegraphics[width=.3\linewidth,trim={16.8cm 4.7cm 17cm 2cm},clip]{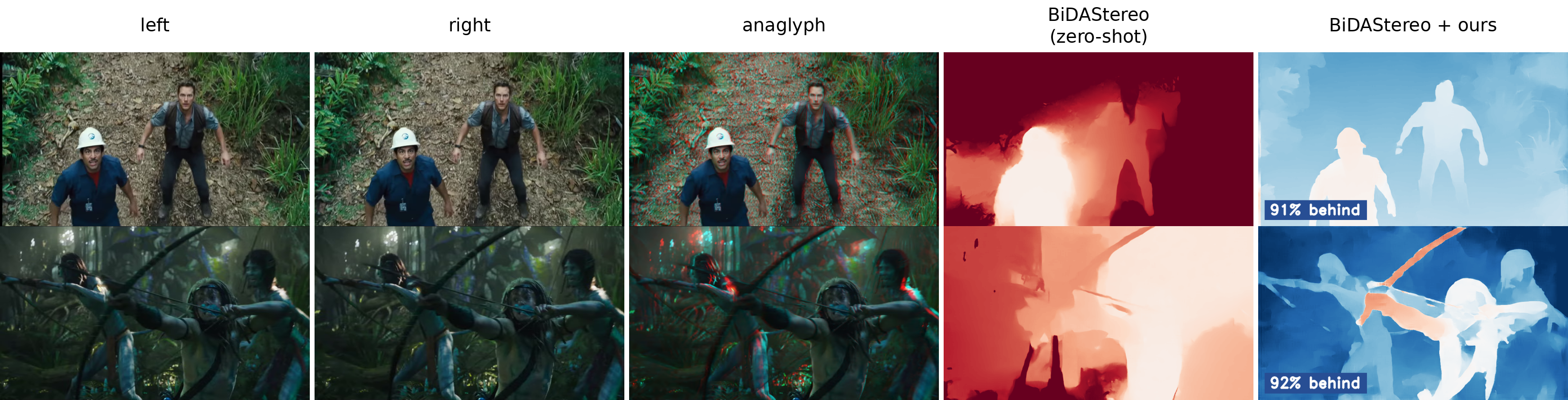}
}
&
\raisebox{-.5\height}{
\includegraphics[width=.3\linewidth,trim={16.8cm 0 17cm 6.7cm},clip]{misc/real3d_movies_bida.png}
}
\end{tabular}
\begin{tabular}{cccc}
    & \makecell{BiDAStereo} 
    & \makecell{DynamicStereo}
    & \makecell{Stereo Any Video}
    \\
    \rotatebox[origin=l]{90}{Zero-shot}
        & \includegraphics[width=.3\linewidth,trim={25.2cm 4.7cm 8.4cm 2cm},clip]{misc/real3d_movies_bida.png}
        & \includegraphics[width=.3\linewidth,trim={25.2cm 4.7cm 8.4cm 2cm},clip]{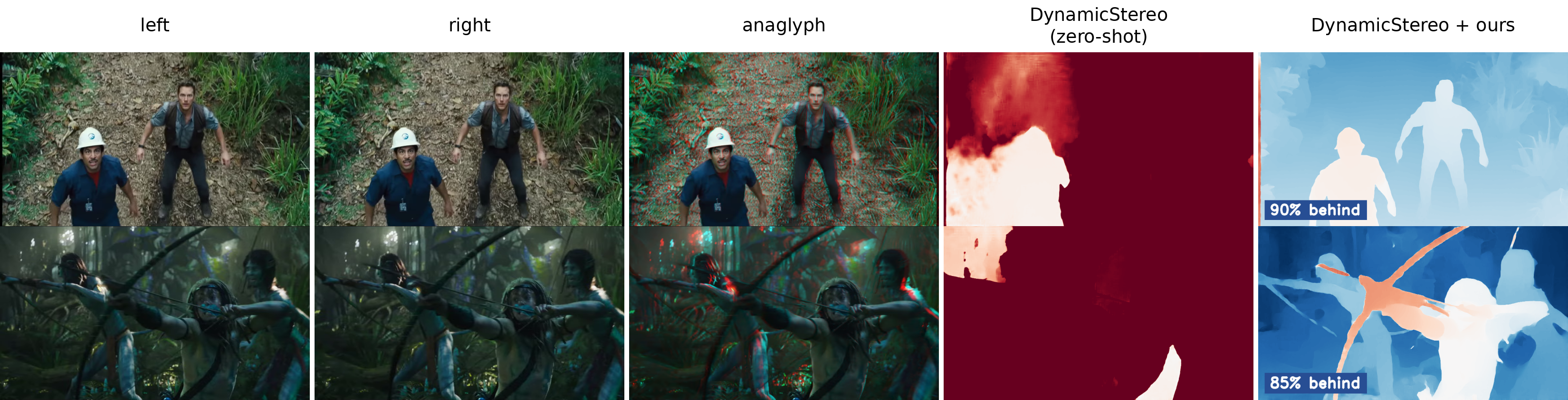}
        & \includegraphics[width=.3\linewidth,trim={25.2cm 4.7cm 8.4cm 2cm},clip]{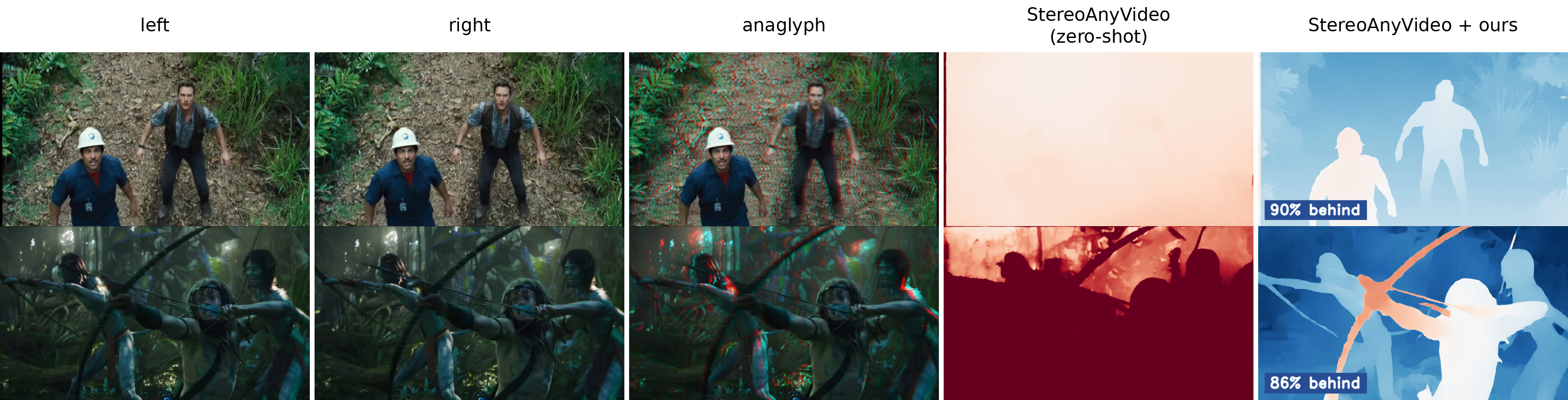}
    \\
    \rotatebox[origin=l]{90}{Ours}
        & \includegraphics[width=.3\linewidth,trim={33.6cm 4.7cm 0 2cm},clip]{misc/real3d_movies_bida.png}
        & \includegraphics[width=.3\linewidth,trim={33.6cm 4.7cm 0 2cm},clip]{misc/real3d_movies_dynamic.png}
        & \includegraphics[width=.3\linewidth,trim={33.6cm 4.7cm 0 2cm},clip]{misc/real3d_movies_sav.png}
    \\
    \rotatebox[origin=l]{90}{Zero-shot}
        & \includegraphics[width=.3\linewidth,trim={25.2cm 0 8.4cm 6.7cm},clip]{misc/real3d_movies_bida.png}
        & \includegraphics[width=.3\linewidth,trim={25.2cm 0 8.4cm 6.7cm},clip]{misc/real3d_movies_dynamic.png}
        & \includegraphics[width=.3\linewidth,trim={25.2cm 0 8.4cm 6.7cm},clip]{misc/real3d_movies_sav.png}
    \\
    \rotatebox[origin=l]{90}{Ours}
        & \includegraphics[width=.3\linewidth,trim={33.6cm 0 0 6.7cm},clip]{misc/real3d_movies_bida.png}
        & \includegraphics[width=.3\linewidth,trim={33.6cm 0 0 6.7cm},clip]{misc/real3d_movies_dynamic.png}
        & \includegraphics[width=.3\linewidth,trim={33.6cm 0 0 6.7cm},clip]{misc/real3d_movies_sav.png}
    \\
    \\
    & \makecell{RAFT-Stereo} 
    & \makecell{IGEV-Stereo}
    & \makecell{FoundationStereo}
    \\
    \rotatebox[origin=l]{90}{Zero-shot}
        & \includegraphics[width=.3\linewidth,trim={25.2cm 4.7cm 8.4cm 2cm},clip]{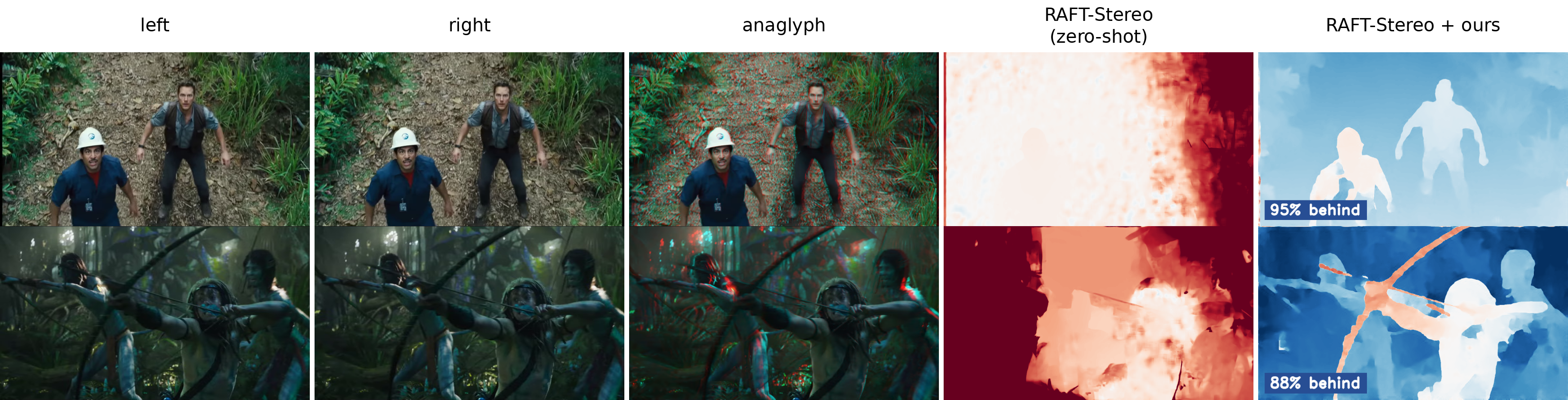}
        & \includegraphics[width=.3\linewidth,trim={25.2cm 4.7cm 8.4cm 2cm},clip]{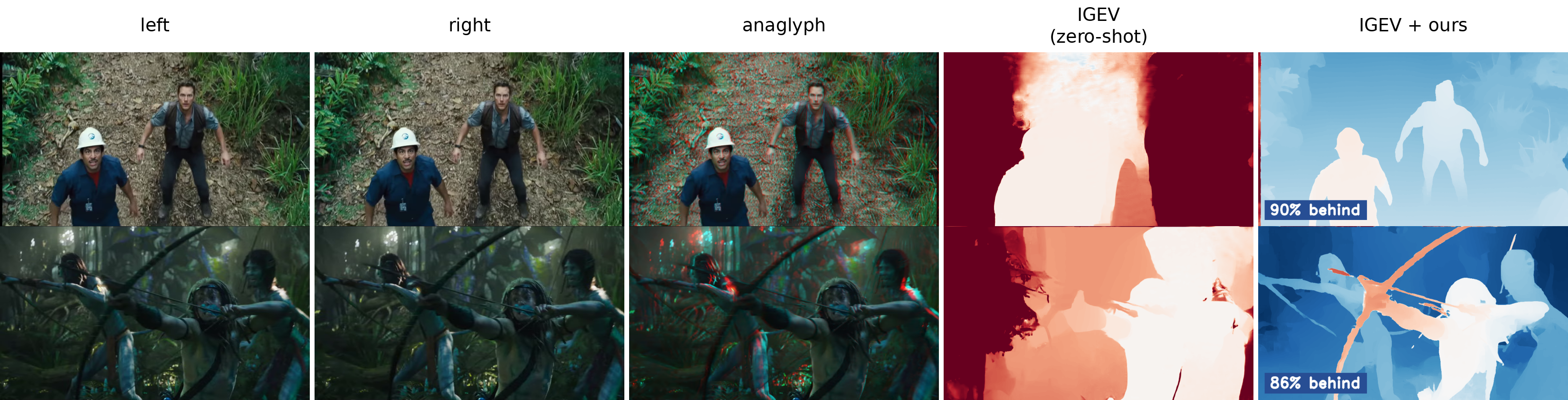}
        & \includegraphics[width=.3\linewidth,trim={25.2cm 4.7cm 8.4cm 2cm},clip]{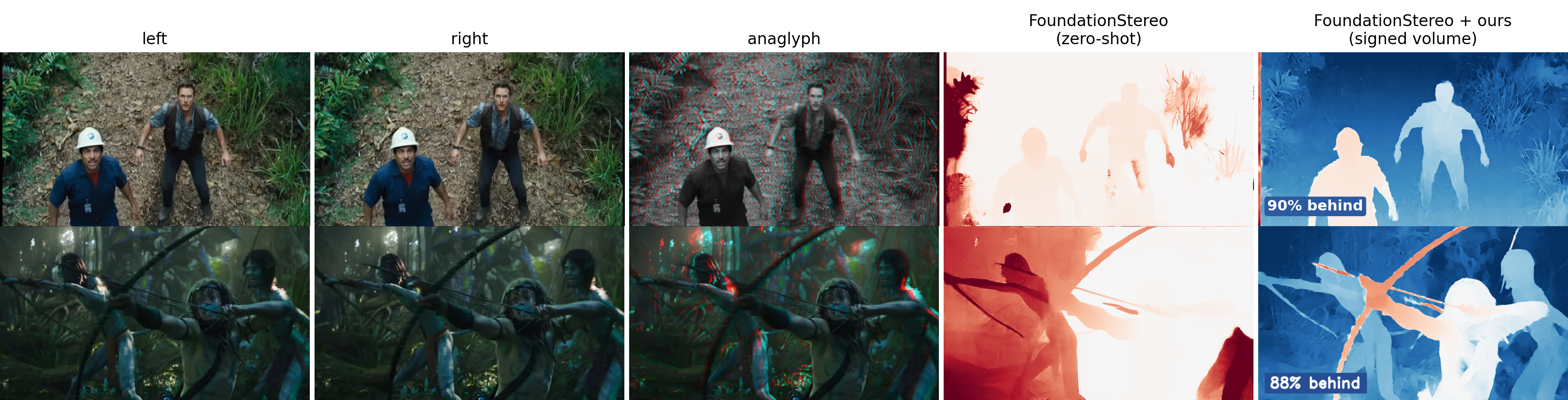}
    \\
    \rotatebox[origin=l]{90}{Ours}
        & \includegraphics[width=.3\linewidth,trim={33.6cm 4.7cm 0 2cm},clip]{misc/real3d_movies_raft.png}
        & \includegraphics[width=.3\linewidth,trim={33.6cm 4.7cm 0 2cm},clip]{misc/real3d_movies_igev.png}
        & \includegraphics[width=.3\linewidth,trim={33.6cm 4.7cm 0 2cm},clip]{misc/real3d_movies.png}
    \\
    \rotatebox[origin=l]{90}{Zero-shot}
        & \includegraphics[width=.3\linewidth,trim={25.2cm 0 8.4cm 6.7cm},clip]{misc/real3d_movies_raft.png}
        & \includegraphics[width=.3\linewidth,trim={25.2cm 0 8.4cm 6.7cm},clip]{misc/real3d_movies_igev.png}
        & \includegraphics[width=.3\linewidth,trim={25.2cm 0 8.4cm 6.7cm},clip]{misc/real3d_movies.png}
    \\
    \rotatebox[origin=l]{90}{Ours}
        & \includegraphics[width=.3\linewidth,trim={33.6cm 0 0 6.7cm},clip]{misc/real3d_movies_raft.png}
        & \includegraphics[width=.3\linewidth,trim={33.6cm 0 0 6.7cm},clip]{misc/real3d_movies_igev.png}
        & \includegraphics[width=.3\linewidth,trim={33.6cm 0 0 6.7cm},clip]{misc/real3d_movies.png}
\end{tabular}
\caption{\textbf{Real-film demonstrations for all the backbones}. Every backbone forces the scene positive on zero-shot evaluations, while our version recovers the signed structure.}
\label{fig:real3d_all}
\end{figure}

\section{Additional Experiments}

\subsection{Non-stereo geometry paradigms under ZDP shift} A natural question is whether the negative-disparity failure simply dissolves under geometry paradigms that never adopt the rectified-stereo formulation. We test one representative of each: VGGT~\citep{wang2025vggt}, which regresses scene-space depth from the two views, and MASt3R~\citep{leroy2024grounding}, whose matching head performs unconstrained 2D correspondence with a sign-symmetric search space. VGGT is evaluated under a deliberately generous protocol: its view-0 depth is converted to disparity by a per-frame \emph{oracle} affine fit in inverse depth, $\hat d = a/\hat Z + b$, least-squares against the signed ground truth (the fit absorbs both the scale ambiguity and the authored shift $\Delta$). MASt3R needs no alignment: reciprocal matches yield signed disparity directly, evaluated on the ${\sim}70\%$ of pixels it matches. As Table~\ref{tab:mvgeo} shows, both are indifferent to the ZDP by construction---EPE is flat across the full signed range, and VGGT's depth for the same frame deviates by only $2.4$--$3.3\%$ across shifts---so neither exhibits the boundary collapse of Table~\ref{tab:main}. Neither, however, approaches the precision the task requires: VGGT sits at ${\sim}17.5$\,px even with oracle alignment, and MASt3R at ${\sim}12$\,px, assigning the correct sign to behind-screen pixels only $35$--$45\%$ of the time. The boundary failure and the sub-pixel precision that motivates repairing it both reside in the rectified-stereo formulation.

\begin{table}[h]
\centering
\caption{Zero-shot non-stereo geometry models on the ZDPShift test split (EPE per $\Delta$ in native px; bad-3 at $\Delta=+32$). VGGT uses the per-frame oracle affine alignment; MASt3R is evaluated on its matched pixels. Both are flat across $\Delta$ but far from the sub-pixel regime of Table~\ref{tab:main}.}
\label{tab:mvgeo}
\small
\begin{tabular}{lrrrrrc}
\toprule
 & \multicolumn{5}{c}{EPE $\downarrow$ @ $\Delta$} & Bad3$\downarrow$ @ ${+}32$ \\
\cmidrule(lr){2-6}\cmidrule(lr){7-7}
Model & $-16$ & $0$ & $+16$ & $+24$ & $+32$ &  \\
\midrule
VGGT (oracle-aligned) & $17.66$ & $17.38$ & $17.57$ & $17.66$ & $17.52$ & $64.2\%$ \\
MASt3R (matched px) & $11.69$ & $11.53$ & $11.50$ & $12.15$ & $12.06$ & $22.6\%$ \\
\bottomrule
\end{tabular}
\end{table}

\subsection{Real-Film Demonstrations Across All Backbones}
Figure~\ref{fig:real3d_all} extends Figure~\ref{fig:real3d} to all the backbones on the same two real-world stereo frames. The pattern of the main paper holds for every backbone: zero-shot, each forces the whole scene into the positive (in-front-of-screen, red) regime, while the version recovers the behind-screen (blue) content, with the drawn bow correctly popping out in front.

\begin{table}[h]
\centering
\caption{\textbf{How much of the network has to adapt.} Each row fine-tunes the same recipe with a different set of parameters left trainable, from a single bias up to the whole network. EPE in native pixels under the protocol of Table~\ref{tab:main}.}
\label{tab:ladder}
\setlength{\tabcolsep}{6pt}
\begin{tabular}{l r r rrrrr r}
\toprule
& Training & & \multicolumn{5}{c}{EPE $\downarrow$ @ $\Delta$} & \\
\cmidrule(lr){4-8}
Trainable set & Params & \% & $-16$ & $0$ & $+16$ & $+24$ & $+32$ & Mean \\
\midrule
\multicolumn{9}{l}{\emph{FoundationStereo}} \\
\cmidrule(lr){1-1}
\quad none (zero-shot)        & $0$          & $0$     & $2.98$ & $2.24$ & $13.06$ & $58.67$ & $75.33$ & $30.46$ \\
\quad final-conv bias         & $1$          & ${<}0.01$ & $4.33$ & $2.96$ & $6.32$ & $9.18$ & $12.50$ & $7.06$ \\
\quad final conv              & $1{,}153$    & $0.003$ & $4.23$ & $2.93$ & $6.17$ & $8.61$ & $10.42$ & $6.47$ \\
\quad disparity head          & $0.43$M      & $1.13$  & $4.17$ & $2.65$ & $4.05$ & $4.66$ & $5.15$ & $4.14$ \\
\quad decoder (head + mask)   & $0.52$M      & $1.38$  & $3.93$ & $2.53$ & $4.22$ & $4.82$ & $5.34$ & $4.17$ \\
\quad update block            & $16.35$M     & $43.5$  & $3.61$ & $2.35$ & $3.39$ & $3.76$ & $3.86$ & $3.39$\\
\quad \;\;+ context encoder   & $22.56$M     & $60.0$  & $3.57$ & $2.21$ & $3.40$ & $3.79$ & $3.70$ & $3.33$ \\
\quad all but feature extractor & $27.59$M   & $73.4$  & $2.89$ & $1.89$ & $2.76$ & $2.83$ & $3.01$ & $2.68$ \\
\quad all (full tuning)       & $37.60$M     & $100$   & $2.97$ & $1.83$ & $2.67$ & $2.78$ & $2.90$ & $\mathbf{2.63}$ \\
\midrule
\multicolumn{9}{l}{\emph{StereoAnyVideo}} \\
\cmidrule(lr){1-1}
\quad none (zero-shot)        & $0$          & $0$     & $3.15$ & $2.73$ & $6.15$ & $9.17$ & $12.91$ & $6.82$ \\
\quad final-conv bias         & $2$          & ${<}0.01$ & $3.17$ & $2.71$ & $6.09$ & $9.11$ & $12.84$ & $6.78$ \\
\quad final conv              & $13{,}826$   & $0.15$  & $4.71$ & $4.44$ & $9.46$ & $11.80$ & $15.86$ & $9.25$ \\
\quad flow head               & $0.90$M      & $9.6$   & $3.89$ & $2.82$ & $3.79$ & $4.49$ & $5.09$ & $4.02$ \\
\quad decoder (head + mask)   & $1.90$M      & $20.2$  & $3.80$ & $2.80$ & $3.64$ & $4.09$ & $4.49$ & $3.76$ \\
\quad \;\;+ GRU + encoder     & $6.11$M      & $65.0$  & $3.44$ & $2.65$ & $3.44$ & $3.90$ & $4.06$ & $3.50$ \\
\quad update block            & $7.09$M      & $75.5$  & $3.44$	& $2.65$ & $3.44$ & $3.76$ & $4.12$ & $3.50$ \\
\quad all but feature encoder & $8.30$M      & $88.3$  & $3.54$ & $2.71$ & $3.48$ & $3.75$ & $4.00$ & $3.50$ \\
\quad all (full tuning)       & $9.40$M      & $100$   & $3.47$ & $2.70$ & $3.35$ & $3.59$ & $3.89$ & $\mathbf{3.40}$ \\
\bottomrule
\end{tabular}
\end{table}

\subsection{Versus the number of adapted parameters}

In this experiment, we further vary the frozen variant training recipe, from a single bias to the whole network (Table~\ref{tab:ladder}).
Capacity saturates early. FoundationStereo reaches $4.14$\,px with $1.13\%$ of itstrainable parameters and StereoAnyVideo $4.02$\,px with $9.6\%$.
Interestingly, one trainable parameter already brings approximately $\sim 4.3\times$ better performance for FoundationStereo, achieving a $7.06$\,px mean EPE across the full disparity spectrum.

\subsection{Can the ZDP simply be manually re-positioned?}
\label{sec:supp_shiftback}

A practitioner could instead re-position the ZDP by shifting the views by hand: reset it to infinity, run a released matcher unchanged, and subtract the offset applied.
That offset, however, cannot be obtained in practice. Delivered stereo does not record it, and recovering it from the images requires a model that already supports negative disparity --- it is a circular problem.
What remains is to assume a maximum budget and shift by that, inflating every disparity in the frame, including in frames that needed no shift at all.
Following the 3DC safety guidelines, the comfortable-viewing budget is $2.9\%$ of image width\footnote{The 3DC safety guidelines put comfortable viewing at a disparity angle of $1$ degree or less. The disparity angle is the difference between the convergence angles at the screen and at the object; for a viewer at distance $D$ with on-screen parallax $p$ it equals $p/D$ independently of interpupillary distance, so $1^\circ$ at the standard viewing distance of three screen heights on a $16{:}9$ display gives $p = 0.029\,W$.}. However, released films may not be bound by it. For example, in a jungle interior from \emph{Avatar: The Way of Water}, peak negative disparity can reach $3.8\%$ of width, exceeding the $2.9\%$ guideline.
Table~\ref{tab:shiftback} reports the comparison under this guideline. Under the only budget a practitioner can apply, our models predict negative disparity without manually assuming ZDP position. As a result, StereoAnyVideo achieves a comparable performance under the ZDP re-position paradigm, while FoundationStereo reaches a significant (over $30\%$) performance gain on the negative regimes.

\begin{table}[h]
\centering
\caption{\textbf{Re-positioning the plane by hand, as an alternative to signed estimation.} EPE in native pixels on the full ZDPShift benchmark. 
\emph{Fixed $2.9\%$} shifts by a constant fraction of image width, the comfortable-viewing budget, which is what a practitioner can apply without knowing the offset.}
\label{tab:shiftback}
\small
\setlength{\tabcolsep}{6pt}
\begin{tabular}{ll | rrrrr}
\toprule
& & \multicolumn{5}{c}{EPE $\downarrow$ @ $\Delta$} \\
\cmidrule(lr){3-7}
Backbone & Condition & $-16$ & $0$ & $+16$ & $+24$ & $+32$ \\
\midrule
\multirow{4}{*}{FoundationStereo}
 & zero-shot                  & $\mathbf{2.91}$ & $2.15$ & $13.03$ & $58.80$ & $75.53$ \\
 & \quad + re-position, fixed $2.9\%$ & $4.35$ & $2.60$ & $3.94$ & $4.48$ & $4.35$ \\
\cmidrule(lr){2-7}
 & {Ours (full tuning)} & $2.97$ & $\mathbf{1.83}$ & $\mathbf{2.67}$ & $\mathbf{2.78}$ & $\mathbf{2.90}$\\
\midrule
\multirow{4}{*}{StereoAnyVideo}
 & zero-shot                  & $\mathbf{3.15}$ & $\mathbf{2.73}$ & $6.15$ & $9.17$ & $12.91$  \\
 & \quad + re-position, fixed $2.9\%$ & $3.70$ & $3.06$ & $3.58$ & $3.79$ & $3.93$  \\
\cmidrule(lr){2-7}
 & {Ours (full tuning)} & ${3.69}$ & $2.75$ & $\mathbf{3.39}$ & $\mathbf{3.66}$ & $\mathbf{3.90}$ \\
\bottomrule
\end{tabular}
\end{table}


\subsection{Ablations}

\paragraph{Data-only ablation of the cost-volume backbones.} Table~\ref{tab:dataonly} reports the ZDPShift data recipe applied to the one-sided cost-volume backbones (IGEV, FoundationStereo) \emph{without} the signed cost volume. The dataset alone helps but leaves a large residual gap in the positive-$\Delta$ (negative-disparity) regime, where the group-wise correlation volume, indexed over $[0, d_{\max}/4]$, is sampled out of bounds and contributes nothing.

\begin{table}[h]
\centering
\caption{\textbf{Data-only ablation of the one-sided cost-volume backbones}, i.e.\ the ZDPShift data recipe without the signed cost volume. EPE (px) per $\Delta$ and bad-3 at $\Delta=+32$. Compare against the signed-volume rows of Table~\ref{tab:main}.}
\label{tab:dataonly}
\small
\begin{tabular}{lrrrrrc}
\toprule
 & \multicolumn{5}{c}{EPE $\downarrow$ @ $\Delta$} & Bad3$\downarrow$ @ $\Delta$ \\
\cmidrule(lr){2-6}\cmidrule(lr){7-7}
Model & $-16$ & $0$ & $+16$ & $+24$ & $+32$ & ${+}32$ \\
\midrule
IGEV \;(data-only)             & $1.03$ & $1.16$ & $2.42$ & $3.33$ & $4.89$  & $8.0\%$ \\
FoundationStereo \;(data-only) & $0.70$ & $0.80$ & $5.09$ & $7.87$ & $11.11$ & $41.1\%$ \\
\bottomrule
\end{tabular}
\end{table}

\paragraph{The signed disparity search closes the gap and unlocks the foundation model.} On IGEV-Stereo, it drops mean EPE to $0.92$\,px, matching the RAFT-Stereo with a per-$\Delta$ profile flat to $0.05$\,px. When applied to FoundationStereo, already the strongest backbone at $\Delta=0$ zero-shot, giving the best result of a mean EPE $0.76$\,px, sub-pixel at every $\Delta$ and bad-3 of $1.8\%$ at $\Delta=+32$. The change does more than restore parity, which lets a foundation model's representation, previously wasted by a one-sided volume, operate across the full signed-disparity range.

\begin{table}[h]
\centering
\caption{Leave-one-out ablation of the recipe ingredients on the in-domain Llamigos test split (same $3$ scenes / $95$ frames per $\Delta$ as Table~\ref{tab:main}). Each row removes one ingredient and reports mean EPE per $\Delta$, and the $\Delta$-mean.}
\label{tab:ablation}
\small
\setlength{\tabcolsep}{5pt}
\begin{tabular}{lrrrrrr}
\toprule
Variant & $\Delta{=}{-}16$ & $\Delta{=}0$ & $\Delta{=}{+}16$ & $\Delta{=}{+}24$ & $\Delta{=}{+}32$ & $\overline{\text{EPE}}$ \\
\midrule
\textbf{Full recipe (multiple $\Delta$ + SceneFlow)} & $\mathbf{0.92}$ & $\mathbf{0.95}$ & $\mathbf{0.97}$ & $\mathbf{0.99}$ & $\mathbf{1.01}$ & $\mathbf{0.97}$ \\
\midrule
$-$R1 ($\Delta=0$ only)     & 1.31 & 0.95 & 6.67 & 12.18 & 15.61 & 7.35 \\
$-$R2 (no SceneFlow rehearsal)& $1.36$ & $1.65$ & $1.60$ & $1.65$ & $1.71$ & $1.59$ \\
\bottomrule
\end{tabular}
\end{table}

\paragraph{Dataset.} Table~\ref{tab:ablation} reports a leave-one-out ablation of the three recipe ingredients, training RAFT-Stereo with each ingredient removed in isolation. SceneFlow rehearsal ($-$R2) accounts for $+0.62$\,px ($+64\%$) of the full recipe's mean EPE, with the cost spread roughly uniformly across all five $\Delta$ shifts --- i.e.\ rehearsal helps the negative-disparity regime as much as the positive one, confirming it acts as a regulariser rather than as a positive-disparity-only safety net.



\subsection{Detailed Performance Breakdown}

The main tables average over the whole benchmark. Tables~\ref{tab:scene_image} and~\ref{tab:scene_video} report the same quantity for every scene and every model at the largest shift $\Delta=+32$, in all three conditions used in the paper: the released weights (\emph{ZS}), full training on HIT (\emph{Full}), and training with the matching features frozen (\emph{Frz}). All numbers follow the protocol of Table~\ref{tab:main}: 960-px inference, errors in native pixels, predictions restricted to the declared search range.

For the released models, what predicts a source's error is simply how much of it sits behind the plane. Project Gold is $90\%$ behind and is the worst source ($711$~px for RAFT-Stereo). Agent 327 is $1\%$ behind and is barely affected ($2.42$~px). The other five fall in between, in order. Nothing about the content of these sources predicts the ordering; only the sign of the disparity does.

Restoring the signed decoder removes that dependence entirely, and turns the ordering around: the sources that were worst become the best, Project Gold ending at $0.75$~px. What is left hardest is Charge and Sprite Fright, and not because of their disparities -- both are among the \emph{least} negative sources. They are the two with smoke, foliage and hair, where the renderer records the geometry behind a semi-transparent element rather than the element itself, so the reference is unreliable. Five such scenes hold half the remaining error, and they are the same five for all six backbones.

Strong variations may also exist within the scenes from the same source movie. Sprite Fright runs from $1.50$ to $15.53$~px. And Agent 327 is the one source that was never restricted in the first place, so it has nothing to regain and only pays the small cost of the change ($2.42\rightarrow3.93$~px for RAFT-Stereo) -- the same trade Table~\ref{tab:external} measures on external benchmarks.

\begin{table}[]
    \centering
\setlength{\tabcolsep}{4.5pt}

\begin{tabular}{l r *{9}{r}}
\toprule
& & \multicolumn{3}{c}{RAFT-Stereo} & \multicolumn{3}{c}{IGEV-Stereo} & \multicolumn{3}{c}{FoundationStereo} \\
\cmidrule(lr){3-5}\cmidrule(lr){6-8}\cmidrule(lr){9-11}
Source / scene & \#fr & ZS & Full & Frz & ZS & Full & Frz & ZS & Full & Frz \\
\midrule
\rowcolor{black!7} \textbf{Agent 327} & \textbf{590} & \textbf{2.42} & \textbf{3.93} & \textbf{3.38} & \textbf{2.25} & \textbf{2.43} & \textbf{2.38} & \textbf{1.61} & \textbf{1.91} & \textbf{1.87} \\
\textcolor{black!55}{\quad \texttt{\scriptsize A327\_02\_01\_A-car\_enter}} & \textcolor{black!55}{94} & \textcolor{black!55}{1.15} & \textcolor{black!55}{4.96} & \textcolor{black!55}{3.48} & \textcolor{black!55}{1.16} & \textcolor{black!55}{1.18} & \textcolor{black!55}{0.70} & \textcolor{black!55}{0.47} & \textcolor{black!55}{0.46} & \textcolor{black!55}{0.47} \\
\textcolor{black!55}{\quad \texttt{\scriptsize A327\_04\_01\_H-sitting}} & \textcolor{black!55}{179} & \textcolor{black!55}{5.05} & \textcolor{black!55}{5.91} & \textcolor{black!55}{5.63} & \textcolor{black!55}{4.95} & \textcolor{black!55}{5.37} & \textcolor{black!55}{5.51} & \textcolor{black!55}{4.04} & \textcolor{black!55}{4.91} & \textcolor{black!55}{4.84} \\
\textcolor{black!55}{\quad \texttt{\scriptsize A327\_07\_04\_F-wall\_slam}} & \textcolor{black!55}{103} & \textcolor{black!55}{0.94} & \textcolor{black!55}{1.45} & \textcolor{black!55}{1.30} & \textcolor{black!55}{0.83} & \textcolor{black!55}{0.86} & \textcolor{black!55}{0.86} & \textcolor{black!55}{0.56} & \textcolor{black!55}{0.58} & \textcolor{black!55}{0.57} \\
\textcolor{black!55}{\quad \texttt{\scriptsize A327\_08\_05\_A-headbutt}} & \textcolor{black!55}{73} & \textcolor{black!55}{2.82} & \textcolor{black!55}{6.11} & \textcolor{black!55}{4.66} & \textcolor{black!55}{2.15} & \textcolor{black!55}{2.31} & \textcolor{black!55}{2.22} & \textcolor{black!55}{1.00} & \textcolor{black!55}{1.15} & \textcolor{black!55}{1.08} \\
\textcolor{black!55}{\quad \texttt{\scriptsize A327\_11\_02\_A-pinned}} & \textcolor{black!55}{52} & \textcolor{black!55}{1.16} & \textcolor{black!55}{2.08} & \textcolor{black!55}{1.73} & \textcolor{black!55}{1.10} & \textcolor{black!55}{1.19} & \textcolor{black!55}{1.11} & \textcolor{black!55}{0.71} & \textcolor{black!55}{0.77} & \textcolor{black!55}{0.72} \\
\textcolor{black!55}{\quad \texttt{\scriptsize A327\_13\_04\_C-return\_of\_th..}} & \textcolor{black!55}{89} & \textcolor{black!55}{0.60} & \textcolor{black!55}{0.99} & \textcolor{black!55}{1.06} & \textcolor{black!55}{0.36} & \textcolor{black!55}{0.44} & \textcolor{black!55}{0.51} & \textcolor{black!55}{0.17} & \textcolor{black!55}{0.22} & \textcolor{black!55}{0.20} \\
\addlinespace[1pt]
\rowcolor{black!7} \textbf{Caminandes} & \textbf{271} & \textbf{13.7} & \textbf{1.77} & \textbf{1.53} & \textbf{10.5} & \textbf{1.19} & \textbf{1.19} & \textbf{14.8} & \textbf{0.83} & \textbf{0.80} \\
\textcolor{black!55}{\quad \texttt{\scriptsize CL\_01\_02\_C.running}} & \textcolor{black!55}{31} & \textcolor{black!55}{23.9} & \textcolor{black!55}{1.61} & \textcolor{black!55}{1.68} & \textcolor{black!55}{26.4} & \textcolor{black!55}{1.69} & \textcolor{black!55}{1.94} & \textcolor{black!55}{25.8} & \textcolor{black!55}{1.52} & \textcolor{black!55}{1.59} \\
\textcolor{black!55}{\quad \texttt{\scriptsize CL\_01\_02\_D.sprinting}} & \textcolor{black!55}{22} & \textcolor{black!55}{17.5} & \textcolor{black!55}{1.27} & \textcolor{black!55}{1.38} & \textcolor{black!55}{18.8} & \textcolor{black!55}{1.24} & \textcolor{black!55}{1.38} & \textcolor{black!55}{17.5} & \textcolor{black!55}{1.27} & \textcolor{black!55}{1.30} \\
\textcolor{black!55}{\quad \texttt{\scriptsize CL\_01\_03\_A.tracks}} & \textcolor{black!55}{42} & \textcolor{black!55}{45.8} & \textcolor{black!55}{0.53} & \textcolor{black!55}{0.52} & \textcolor{black!55}{23.0} & \textcolor{black!55}{0.42} & \textcolor{black!55}{0.49} & \textcolor{black!55}{55.2} & \textcolor{black!55}{0.29} & \textcolor{black!55}{0.37} \\
\textcolor{black!55}{\quad \texttt{\scriptsize CL\_10\_01\_A.supper}} & \textcolor{black!55}{176} & \textcolor{black!55}{3.76} & \textcolor{black!55}{2.16} & \textcolor{black!55}{1.77} & \textcolor{black!55}{3.73} & \textcolor{black!55}{1.28} & \textcolor{black!55}{1.21} & \textcolor{black!55}{2.97} & \textcolor{black!55}{0.78} & \textcolor{black!55}{0.71} \\
\addlinespace[1pt]
\rowcolor{black!7} \textbf{Charge} & \textbf{435} & \textbf{74.2} & \textbf{11.0} & \textbf{12.1} & \textbf{65.6} & \textbf{8.74} & \textbf{9.08} & \textbf{85.9} & \textbf{5.81} & \textbf{6.82} \\
\textcolor{black!55}{\quad \texttt{\scriptsize Ch\_020\_0020}} & \textcolor{black!55}{84} & \textcolor{black!55}{103} & \textcolor{black!55}{19.8} & \textcolor{black!55}{20.1} & \textcolor{black!55}{145} & \textcolor{black!55}{15.8} & \textcolor{black!55}{13.6} & \textcolor{black!55}{306} & \textcolor{black!55}{12.0} & \textcolor{black!55}{16.7} \\
\textcolor{black!55}{\quad \texttt{\scriptsize Ch\_040\_0040}} & \textcolor{black!55}{164} & \textcolor{black!55}{127} & \textcolor{black!55}{8.73} & \textcolor{black!55}{10.9} & \textcolor{black!55}{86.1} & \textcolor{black!55}{6.86} & \textcolor{black!55}{8.38} & \textcolor{black!55}{57.7} & \textcolor{black!55}{3.12} & \textcolor{black!55}{2.89} \\
\textcolor{black!55}{\quad \texttt{\scriptsize Ch\_050\_0160}} & \textcolor{black!55}{70} & \textcolor{black!55}{17.6} & \textcolor{black!55}{13.2} & \textcolor{black!55}{13.4} & \textcolor{black!55}{14.8} & \textcolor{black!55}{11.5} & \textcolor{black!55}{12.7} & \textcolor{black!55}{15.1} & \textcolor{black!55}{7.33} & \textcolor{black!55}{8.85} \\
\textcolor{black!55}{\quad \texttt{\scriptsize Ch\_060\_0100}} & \textcolor{black!55}{42} & \textcolor{black!55}{22.8} & \textcolor{black!55}{4.13} & \textcolor{black!55}{5.16} & \textcolor{black!55}{14.3} & \textcolor{black!55}{2.54} & \textcolor{black!55}{2.46} & \textcolor{black!55}{12.7} & \textcolor{black!55}{1.94} & \textcolor{black!55}{1.69} \\
\textcolor{black!55}{\quad \texttt{\scriptsize Ch\_060\_0130}} & \textcolor{black!55}{75} & \textcolor{black!55}{8.30} & \textcolor{black!55}{7.92} & \textcolor{black!55}{8.29} & \textcolor{black!55}{8.19} & \textcolor{black!55}{5.82} & \textcolor{black!55}{5.89} & \textcolor{black!55}{7.57} & \textcolor{black!55}{5.44} & \textcolor{black!55}{5.31} \\
\addlinespace[1pt]
\rowcolor{black!7} \textbf{Project Gold} & \textbf{226} & \textbf{711} & \textbf{0.75} & \textbf{0.66} & \textbf{33.7} & \textbf{0.37} & \textbf{0.37} & \textbf{149} & \textbf{0.39} & \textbf{0.32} \\
\textcolor{black!55}{\quad \texttt{\scriptsize PG\_265\_0010}} & \textcolor{black!55}{226} & \textcolor{black!55}{711} & \textcolor{black!55}{0.75} & \textcolor{black!55}{0.66} & \textcolor{black!55}{33.7} & \textcolor{black!55}{0.37} & \textcolor{black!55}{0.37} & \textcolor{black!55}{149} & \textcolor{black!55}{0.39} & \textcolor{black!55}{0.32} \\
\addlinespace[1pt]
\rowcolor{black!7} \textbf{Settlers} & \textbf{1283} & \textbf{317} & \textbf{1.98} & \textbf{2.71} & \textbf{82.1} & \textbf{1.83} & \textbf{1.99} & \textbf{167} & \textbf{1.62} & \textbf{1.51} \\
\textcolor{black!55}{\quad \texttt{\scriptsize St\_01-desert}} & \textcolor{black!55}{204} & \textcolor{black!55}{21.1} & \textcolor{black!55}{2.37} & \textcolor{black!55}{2.71} & \textcolor{black!55}{24.9} & \textcolor{black!55}{2.01} & \textcolor{black!55}{2.06} & \textcolor{black!55}{31.8} & \textcolor{black!55}{1.46} & \textcolor{black!55}{1.74} \\
\textcolor{black!55}{\quad \texttt{\scriptsize St\_02-phileas}} & \textcolor{black!55}{260} & \textcolor{black!55}{12.2} & \textcolor{black!55}{3.35} & \textcolor{black!55}{5.39} & \textcolor{black!55}{11.7} & \textcolor{black!55}{3.06} & \textcolor{black!55}{3.58} & \textcolor{black!55}{13.0} & \textcolor{black!55}{2.29} & \textcolor{black!55}{2.49} \\
\textcolor{black!55}{\quad \texttt{\scriptsize St\_gabby\_anim}} & \textcolor{black!55}{376} & \textcolor{black!55}{12.9} & \textcolor{black!55}{1.28} & \textcolor{black!55}{1.78} & \textcolor{black!55}{12.9} & \textcolor{black!55}{0.84} & \textcolor{black!55}{0.82} & \textcolor{black!55}{12.0} & \textcolor{black!55}{0.62} & \textcolor{black!55}{0.50} \\
\textcolor{black!55}{\quad \texttt{\scriptsize St\_pip\_anim}} & \textcolor{black!55}{443} & \textcolor{black!55}{891} & \textcolor{black!55}{1.59} & \textcolor{black!55}{1.94} & \textcolor{black!55}{208} & \textcolor{black!55}{1.87} & \textcolor{black!55}{2.02} & \textcolor{black!55}{451} & \textcolor{black!55}{2.16} & \textcolor{black!55}{1.68} \\
\addlinespace[1pt]
\rowcolor{black!7} \textbf{Spring} & \textbf{649} & \textbf{28.2} & \textbf{2.43} & \textbf{2.18} & \textbf{39.9} & \textbf{2.33} & \textbf{2.33} & \textbf{23.8} & \textbf{1.28} & \textbf{1.26} \\
\textcolor{black!55}{\quad \texttt{\scriptsize Sp\_01\_025\_A}} & \textcolor{black!55}{128} & \textcolor{black!55}{5.51} & \textcolor{black!55}{7.73} & \textcolor{black!55}{6.26} & \textcolor{black!55}{4.55} & \textcolor{black!55}{7.76} & \textcolor{black!55}{7.65} & \textcolor{black!55}{3.05} & \textcolor{black!55}{3.15} & \textcolor{black!55}{3.15} \\
\textcolor{black!55}{\quad \texttt{\scriptsize Sp\_02\_055\_A}} & \textcolor{black!55}{263} & \textcolor{black!55}{1.02} & \textcolor{black!55}{0.95} & \textcolor{black!55}{0.93} & \textcolor{black!55}{1.07} & \textcolor{black!55}{0.80} & \textcolor{black!55}{0.78} & \textcolor{black!55}{0.67} & \textcolor{black!55}{0.62} & \textcolor{black!55}{0.61} \\
\textcolor{black!55}{\quad \texttt{\scriptsize Sp\_06\_005\_A}} & \textcolor{black!55}{44} & \textcolor{black!55}{57.9} & \textcolor{black!55}{1.28} & \textcolor{black!55}{1.39} & \textcolor{black!55}{83.3} & \textcolor{black!55}{1.10} & \textcolor{black!55}{1.19} & \textcolor{black!55}{113} & \textcolor{black!55}{1.03} & \textcolor{black!55}{1.02} \\
\textcolor{black!55}{\quad \texttt{\scriptsize Sp\_06\_035\_A}} & \textcolor{black!55}{117} & \textcolor{black!55}{17.7} & \textcolor{black!55}{1.48} & \textcolor{black!55}{1.65} & \textcolor{black!55}{33.3} & \textcolor{black!55}{1.37} & \textcolor{black!55}{1.48} & \textcolor{black!55}{31.1} & \textcolor{black!55}{1.25} & \textcolor{black!55}{1.17} \\
\textcolor{black!55}{\quad \texttt{\scriptsize Sp\_10\_020\_A}} & \textcolor{black!55}{97} & \textcolor{black!55}{131} & \textcolor{black!55}{1.08} & \textcolor{black!55}{1.15} & \textcolor{black!55}{180} & \textcolor{black!55}{1.00} & \textcolor{black!55}{1.03} & \textcolor{black!55}{64.9} & \textcolor{black!55}{0.75} & \textcolor{black!55}{0.73} \\
\addlinespace[1pt]
\rowcolor{black!7} \textbf{Sprite Fright} & \textbf{821} & \textbf{23.9} & \textbf{9.36} & \textbf{9.42} & \textbf{24.2} & \textbf{7.86} & \textbf{8.05} & \textbf{21.2} & \textbf{6.42} & \textbf{6.63} \\
\textcolor{black!55}{\quad \texttt{\scriptsize SF\_020\_0060\_A}} & \textcolor{black!55}{48} & \textcolor{black!55}{35.3} & \textcolor{black!55}{10.7} & \textcolor{black!55}{10.9} & \textcolor{black!55}{35.6} & \textcolor{black!55}{9.99} & \textcolor{black!55}{10.1} & \textcolor{black!55}{66.2} & \textcolor{black!55}{10.3} & \textcolor{black!55}{10.2} \\
\textcolor{black!55}{\quad \texttt{\scriptsize SF\_030\_0020\_A}} & \textcolor{black!55}{70} & \textcolor{black!55}{21.8} & \textcolor{black!55}{2.46} & \textcolor{black!55}{2.49} & \textcolor{black!55}{43.3} & \textcolor{black!55}{1.74} & \textcolor{black!55}{1.34} & \textcolor{black!55}{53.5} & \textcolor{black!55}{1.63} & \textcolor{black!55}{1.32} \\
\textcolor{black!55}{\quad \texttt{\scriptsize SF\_030\_0060\_A}} & \textcolor{black!55}{126} & \textcolor{black!55}{10.4} & \textcolor{black!55}{10.8} & \textcolor{black!55}{10.6} & \textcolor{black!55}{10.2} & \textcolor{black!55}{8.26} & \textcolor{black!55}{8.65} & \textcolor{black!55}{7.37} & \textcolor{black!55}{7.15} & \textcolor{black!55}{7.09} \\
\textcolor{black!55}{\quad \texttt{\scriptsize SF\_030\_0070\_A}} & \textcolor{black!55}{291} & \textcolor{black!55}{35.6} & \textcolor{black!55}{6.69} & \textcolor{black!55}{6.65} & \textcolor{black!55}{32.0} & \textcolor{black!55}{5.70} & \textcolor{black!55}{4.91} & \textcolor{black!55}{22.3} & \textcolor{black!55}{3.48} & \textcolor{black!55}{3.82} \\
\textcolor{black!55}{\quad \texttt{\scriptsize SF\_050\_0150\_A}} & \textcolor{black!55}{160} & \textcolor{black!55}{25.1} & \textcolor{black!55}{20.9} & \textcolor{black!55}{21.1} & \textcolor{black!55}{22.9} & \textcolor{black!55}{17.8} & \textcolor{black!55}{20.0} & \textcolor{black!55}{15.4} & \textcolor{black!55}{15.5} & \textcolor{black!55}{15.5} \\
\textcolor{black!55}{\quad \texttt{\scriptsize SF\_080\_0010\_A}} & \textcolor{black!55}{86} & \textcolor{black!55}{7.80} & \textcolor{black!55}{3.15} & \textcolor{black!55}{3.28} & \textcolor{black!55}{8.93} & \textcolor{black!55}{2.79} & \textcolor{black!55}{2.80} & \textcolor{black!55}{6.70} & \textcolor{black!55}{2.41} & \textcolor{black!55}{2.22} \\
\textcolor{black!55}{\quad \texttt{\scriptsize SF\_110\_0180\_A}} & \textcolor{black!55}{40} & \textcolor{black!55}{1.26} & \textcolor{black!55}{2.18} & \textcolor{black!55}{2.52} & \textcolor{black!55}{2.05} & \textcolor{black!55}{1.78} & \textcolor{black!55}{1.95} & \textcolor{black!55}{1.10} & \textcolor{black!55}{1.50} & \textcolor{black!55}{4.68} \\
\midrule
\textbf{All scenes} & \textbf{4275} & \textbf{150} & \textbf{4.57} & \textbf{4.78} & \textbf{44.8} & \textbf{3.73} & \textbf{3.85} & \textbf{75.5} & \textbf{2.84} & \textbf{2.94} \\
\bottomrule
\end{tabular}

    \caption{\textbf{Source- and scene-level results on ZDPShift benchmark, with image matchers ($\Delta=+32$).}}
    \label{tab:scene_image}
\end{table}

\begin{table}[]
    \centering
\setlength{\tabcolsep}{4.5pt}
\begin{tabular}{l r *{9}{r}}
\toprule
& & \multicolumn{3}{c}{DynamicStereo} & \multicolumn{3}{c}{BiDAStereo} & \multicolumn{3}{c}{StereoAnyVideo} \\
\cmidrule(lr){3-5}\cmidrule(lr){6-8}\cmidrule(lr){9-11}
Source / scene & \#fr & ZS & Full & Frz & ZS & Full & Frz & ZS & Full & Frz \\
\midrule
\rowcolor{black!7} \textbf{Agent 327} & \textbf{590} & \textbf{1.63} & \textbf{2.10} & \textbf{1.69} & \textbf{1.68} & \textbf{2.18} & \textbf{1.92} & \textbf{1.39} & \textbf{1.57} & \textbf{1.54} \\
\textcolor{black!55}{\quad \texttt{\scriptsize A327\_02\_01\_A-car\_enter}} & \textcolor{black!55}{94} & \textcolor{black!55}{0.93} & \textcolor{black!55}{2.25} & \textcolor{black!55}{1.23} & \textcolor{black!55}{1.12} & \textcolor{black!55}{1.28} & \textcolor{black!55}{1.00} & \textcolor{black!55}{0.67} & \textcolor{black!55}{0.74} & \textcolor{black!55}{0.73} \\
\textcolor{black!55}{\quad \texttt{\scriptsize A327\_04\_01\_H-sitting}} & \textcolor{black!55}{179} & \textcolor{black!55}{4.25} & \textcolor{black!55}{4.21} & \textcolor{black!55}{4.10} & \textcolor{black!55}{4.22} & \textcolor{black!55}{4.48} & \textcolor{black!55}{4.56} & \textcolor{black!55}{4.12} & \textcolor{black!55}{4.40} & \textcolor{black!55}{4.35} \\
\textcolor{black!55}{\quad \texttt{\scriptsize A327\_07\_04\_F-wall\_slam}} & \textcolor{black!55}{103} & \textcolor{black!55}{0.54} & \textcolor{black!55}{0.68} & \textcolor{black!55}{0.53} & \textcolor{black!55}{0.58} & \textcolor{black!55}{0.73} & \textcolor{black!55}{0.61} & \textcolor{black!55}{0.46} & \textcolor{black!55}{0.57} & \textcolor{black!55}{0.50} \\
\textcolor{black!55}{\quad \texttt{\scriptsize A327\_08\_05\_A-headbutt}} & \textcolor{black!55}{73} & \textcolor{black!55}{2.44} & \textcolor{black!55}{3.40} & \textcolor{black!55}{2.53} & \textcolor{black!55}{2.61} & \textcolor{black!55}{4.81} & \textcolor{black!55}{3.75} & \textcolor{black!55}{1.59} & \textcolor{black!55}{2.09} & \textcolor{black!55}{2.09} \\
\textcolor{black!55}{\quad \texttt{\scriptsize A327\_11\_02\_A-pinned}} & \textcolor{black!55}{52} & \textcolor{black!55}{1.15} & \textcolor{black!55}{1.34} & \textcolor{black!55}{1.09} & \textcolor{black!55}{1.18} & \textcolor{black!55}{1.29} & \textcolor{black!55}{1.11} & \textcolor{black!55}{1.23} & \textcolor{black!55}{1.21} & \textcolor{black!55}{1.14} \\
\textcolor{black!55}{\quad \texttt{\scriptsize A327\_13\_04\_C-return\_of\_th..}} & \textcolor{black!55}{89} & \textcolor{black!55}{0.47} & \textcolor{black!55}{0.74} & \textcolor{black!55}{0.68} & \textcolor{black!55}{0.35} & \textcolor{black!55}{0.48} & \textcolor{black!55}{0.50} & \textcolor{black!55}{0.26} & \textcolor{black!55}{0.43} & \textcolor{black!55}{0.41} \\
\addlinespace[1pt]
\rowcolor{black!7} \textbf{Caminandes} & \textbf{271} & \textbf{20.8} & \textbf{1.41} & \textbf{1.25} & \textbf{16.1} & \textbf{1.31} & \textbf{1.16} & \textbf{11.8} & \textbf{1.09} & \textbf{1.05} \\
\textcolor{black!55}{\quad \texttt{\scriptsize CL\_01\_02\_C.running}} & \textcolor{black!55}{31} & \textcolor{black!55}{56.2} & \textcolor{black!55}{1.76} & \textcolor{black!55}{1.81} & \textcolor{black!55}{39.4} & \textcolor{black!55}{1.89} & \textcolor{black!55}{1.93} & \textcolor{black!55}{25.5} & \textcolor{black!55}{1.73} & \textcolor{black!55}{1.75} \\
\textcolor{black!55}{\quad \texttt{\scriptsize CL\_01\_02\_D.sprinting}} & \textcolor{black!55}{22} & \textcolor{black!55}{20.3} & \textcolor{black!55}{1.40} & \textcolor{black!55}{1.28} & \textcolor{black!55}{18.8} & \textcolor{black!55}{1.23} & \textcolor{black!55}{1.21} & \textcolor{black!55}{16.5} & \textcolor{black!55}{1.17} & \textcolor{black!55}{1.14} \\
\textcolor{black!55}{\quad \texttt{\scriptsize CL\_01\_03\_A.tracks}} & \textcolor{black!55}{42} & \textcolor{black!55}{2.53} & \textcolor{black!55}{0.35} & \textcolor{black!55}{0.80} & \textcolor{black!55}{1.86} & \textcolor{black!55}{0.45} & \textcolor{black!55}{0.32} & \textcolor{black!55}{1.46} & \textcolor{black!55}{0.26} & \textcolor{black!55}{0.25} \\
\textcolor{black!55}{\quad \texttt{\scriptsize CL\_10\_01\_A.supper}} & \textcolor{black!55}{176} & \textcolor{black!55}{4.06} & \textcolor{black!55}{2.13} & \textcolor{black!55}{1.12} & \textcolor{black!55}{4.33} & \textcolor{black!55}{1.65} & \textcolor{black!55}{1.17} & \textcolor{black!55}{3.71} & \textcolor{black!55}{1.22} & \textcolor{black!55}{1.06} \\
\addlinespace[1pt]
\rowcolor{black!7} \textbf{Charge} & \textbf{435} & \textbf{31.0} & \textbf{9.75} & \textbf{10.0} & \textbf{43.4} & \textbf{9.88} & \textbf{10.2} & \textbf{20.3} & \textbf{9.40} & \textbf{8.97} \\
\textcolor{black!55}{\quad \texttt{\scriptsize Ch\_020\_0020}} & \textcolor{black!55}{84} & \textcolor{black!55}{23.3} & \textcolor{black!55}{18.1} & \textcolor{black!55}{19.5} & \textcolor{black!55}{59.1} & \textcolor{black!55}{19.1} & \textcolor{black!55}{19.5} & \textcolor{black!55}{26.7} & \textcolor{black!55}{19.6} & \textcolor{black!55}{20.1} \\
\textcolor{black!55}{\quad \texttt{\scriptsize Ch\_040\_0040}} & \textcolor{black!55}{164} & \textcolor{black!55}{92.7} & \textcolor{black!55}{6.34} & \textcolor{black!55}{7.23} & \textcolor{black!55}{105} & \textcolor{black!55}{7.99} & \textcolor{black!55}{10.1} & \textcolor{black!55}{39.7} & \textcolor{black!55}{4.96} & \textcolor{black!55}{5.10} \\
\textcolor{black!55}{\quad \texttt{\scriptsize Ch\_050\_0160}} & \textcolor{black!55}{70} & \textcolor{black!55}{15.2} & \textcolor{black!55}{12.1} & \textcolor{black!55}{12.3} & \textcolor{black!55}{18.8} & \textcolor{black!55}{12.7} & \textcolor{black!55}{12.5} & \textcolor{black!55}{16.6} & \textcolor{black!55}{10.7} & \textcolor{black!55}{10.9} \\
\textcolor{black!55}{\quad \texttt{\scriptsize Ch\_060\_0100}} & \textcolor{black!55}{42} & \textcolor{black!55}{15.6} & \textcolor{black!55}{6.05} & \textcolor{black!55}{5.82} & \textcolor{black!55}{26.1} & \textcolor{black!55}{4.61} & \textcolor{black!55}{3.90} & \textcolor{black!55}{12.2} & \textcolor{black!55}{5.83} & \textcolor{black!55}{3.81} \\
\textcolor{black!55}{\quad \texttt{\scriptsize Ch\_060\_0130}} & \textcolor{black!55}{75} & \textcolor{black!55}{7.95} & \textcolor{black!55}{6.25} & \textcolor{black!55}{5.14} & \textcolor{black!55}{8.33} & \textcolor{black!55}{5.00} & \textcolor{black!55}{5.11} & \textcolor{black!55}{6.52} & \textcolor{black!55}{5.93} & \textcolor{black!55}{4.93} \\
\addlinespace[1pt]
\rowcolor{black!7} \textbf{Project Gold} & \textbf{226} & \textbf{182} & \textbf{0.38} & \textbf{0.32} & \textbf{36.2} & \textbf{0.40} & \textbf{0.41} & \textbf{38.1} & \textbf{0.23} & \textbf{0.23} \\
\textcolor{black!55}{\quad \texttt{\scriptsize PG\_265\_0010}} & \textcolor{black!55}{226} & \textcolor{black!55}{182} & \textcolor{black!55}{0.38} & \textcolor{black!55}{0.32} & \textcolor{black!55}{36.2} & \textcolor{black!55}{0.40} & \textcolor{black!55}{0.41} & \textcolor{black!55}{38.1} & \textcolor{black!55}{0.23} & \textcolor{black!55}{0.23} \\
\addlinespace[1pt]
\rowcolor{black!7} \textbf{Settlers} & \textbf{1283} & \textbf{16.0} & \textbf{3.06} & \textbf{3.37} & \textbf{124} & \textbf{2.58} & \textbf{2.71} & \textbf{20.3} & \textbf{1.92} & \textbf{2.34} \\
\textcolor{black!55}{\quad \texttt{\scriptsize St\_01-desert}} & \textcolor{black!55}{204} & \textcolor{black!55}{20.3} & \textcolor{black!55}{2.71} & \textcolor{black!55}{3.97} & \textcolor{black!55}{28.3} & \textcolor{black!55}{3.30} & \textcolor{black!55}{3.65} & \textcolor{black!55}{17.2} & \textcolor{black!55}{2.29} & \textcolor{black!55}{2.95} \\
\textcolor{black!55}{\quad \texttt{\scriptsize St\_02-phileas}} & \textcolor{black!55}{260} & \textcolor{black!55}{12.7} & \textcolor{black!55}{7.01} & \textcolor{black!55}{6.49} & \textcolor{black!55}{14.1} & \textcolor{black!55}{4.11} & \textcolor{black!55}{3.91} & \textcolor{black!55}{11.8} & \textcolor{black!55}{3.33} & \textcolor{black!55}{3.87} \\
\textcolor{black!55}{\quad \texttt{\scriptsize St\_gabby\_anim}} & \textcolor{black!55}{376} & \textcolor{black!55}{15.9} & \textcolor{black!55}{0.99} & \textcolor{black!55}{1.11} & \textcolor{black!55}{17.7} & \textcolor{black!55}{1.25} & \textcolor{black!55}{1.53} & \textcolor{black!55}{12.4} & \textcolor{black!55}{0.82} & \textcolor{black!55}{0.92} \\
\textcolor{black!55}{\quad \texttt{\scriptsize St\_pip\_anim}} & \textcolor{black!55}{443} & \textcolor{black!55}{15.1} & \textcolor{black!55}{1.53} & \textcolor{black!55}{1.89} & \textcolor{black!55}{437} & \textcolor{black!55}{1.64} & \textcolor{black!55}{1.74} & \textcolor{black!55}{39.6} & \textcolor{black!55}{1.24} & \textcolor{black!55}{1.62} \\
\addlinespace[1pt]
\rowcolor{black!7} \textbf{Spring} & \textbf{649} & \textbf{15.0} & \textbf{2.08} & \textbf{2.07} & \textbf{34.8} & \textbf{1.98} & \textbf{1.93} & \textbf{11.4} & \textbf{1.93} & \textbf{2.12} \\
\textcolor{black!55}{\quad \texttt{\scriptsize Sp\_01\_025\_A}} & \textcolor{black!55}{128} & \textcolor{black!55}{5.14} & \textcolor{black!55}{5.35} & \textcolor{black!55}{5.17} & \textcolor{black!55}{4.74} & \textcolor{black!55}{4.82} & \textcolor{black!55}{4.63} & \textcolor{black!55}{3.90} & \textcolor{black!55}{5.10} & \textcolor{black!55}{5.78} \\
\textcolor{black!55}{\quad \texttt{\scriptsize Sp\_02\_055\_A}} & \textcolor{black!55}{263} & \textcolor{black!55}{1.06} & \textcolor{black!55}{0.73} & \textcolor{black!55}{0.73} & \textcolor{black!55}{0.85} & \textcolor{black!55}{0.71} & \textcolor{black!55}{0.71} & \textcolor{black!55}{1.02} & \textcolor{black!55}{0.62} & \textcolor{black!55}{0.62} \\
\textcolor{black!55}{\quad \texttt{\scriptsize Sp\_06\_005\_A}} & \textcolor{black!55}{44} & \textcolor{black!55}{22.0} & \textcolor{black!55}{2.16} & \textcolor{black!55}{2.06} & \textcolor{black!55}{39.3} & \textcolor{black!55}{2.25} & \textcolor{black!55}{2.17} & \textcolor{black!55}{23.7} & \textcolor{black!55}{2.01} & \textcolor{black!55}{2.23} \\
\textcolor{black!55}{\quad \texttt{\scriptsize Sp\_06\_035\_A}} & \textcolor{black!55}{117} & \textcolor{black!55}{20.9} & \textcolor{black!55}{0.99} & \textcolor{black!55}{1.16} & \textcolor{black!55}{33.6} & \textcolor{black!55}{1.00} & \textcolor{black!55}{1.00} & \textcolor{black!55}{12.0} & \textcolor{black!55}{0.89} & \textcolor{black!55}{0.91} \\
\textcolor{black!55}{\quad \texttt{\scriptsize Sp\_10\_020\_A}} & \textcolor{black!55}{97} & \textcolor{black!55}{25.9} & \textcolor{black!55}{1.19} & \textcolor{black!55}{1.25} & \textcolor{black!55}{95.3} & \textcolor{black!55}{1.10} & \textcolor{black!55}{1.12} & \textcolor{black!55}{16.2} & \textcolor{black!55}{1.04} & \textcolor{black!55}{1.05} \\
\addlinespace[1pt]
\rowcolor{black!7} \textbf{Sprite Fright} & \textbf{821} & \textbf{13.7} & \textbf{10.2} & \textbf{10.3} & \textbf{24.2} & \textbf{7.41} & \textbf{7.38} & \textbf{11.5} & \textbf{6.68} & \textbf{7.11} \\
\textcolor{black!55}{\quad \texttt{\scriptsize SF\_020\_0060\_A}} & \textcolor{black!55}{48} & \textcolor{black!55}{24.3} & \textcolor{black!55}{10.3} & \textcolor{black!55}{10.6} & \textcolor{black!55}{33.9} & \textcolor{black!55}{10.6} & \textcolor{black!55}{10.4} & \textcolor{black!55}{21.1} & \textcolor{black!55}{9.62} & \textcolor{black!55}{10.1} \\
\textcolor{black!55}{\quad \texttt{\scriptsize SF\_030\_0020\_A}} & \textcolor{black!55}{70} & \textcolor{black!55}{10.5} & \textcolor{black!55}{2.29} & \textcolor{black!55}{2.01} & \textcolor{black!55}{31.4} & \textcolor{black!55}{2.30} & \textcolor{black!55}{2.34} & \textcolor{black!55}{8.85} & \textcolor{black!55}{1.77} & \textcolor{black!55}{1.87} \\
\textcolor{black!55}{\quad \texttt{\scriptsize SF\_030\_0060\_A}} & \textcolor{black!55}{126} & \textcolor{black!55}{9.80} & \textcolor{black!55}{10.7} & \textcolor{black!55}{10.6} & \textcolor{black!55}{9.37} & \textcolor{black!55}{9.76} & \textcolor{black!55}{9.69} & \textcolor{black!55}{8.08} & \textcolor{black!55}{9.99} & \textcolor{black!55}{10.3} \\
\textcolor{black!55}{\quad \texttt{\scriptsize SF\_030\_0070\_A}} & \textcolor{black!55}{291} & \textcolor{black!55}{20.6} & \textcolor{black!55}{6.68} & \textcolor{black!55}{6.18} & \textcolor{black!55}{56.8} & \textcolor{black!55}{6.61} & \textcolor{black!55}{6.69} & \textcolor{black!55}{13.9} & \textcolor{black!55}{5.40} & \textcolor{black!55}{4.87} \\
\textcolor{black!55}{\quad \texttt{\scriptsize SF\_050\_0150\_A}} & \textcolor{black!55}{160} & \textcolor{black!55}{20.3} & \textcolor{black!55}{18.5} & \textcolor{black!55}{17.3} & \textcolor{black!55}{21.6} & \textcolor{black!55}{18.2} & \textcolor{black!55}{18.2} & \textcolor{black!55}{18.2} & \textcolor{black!55}{15.8} & \textcolor{black!55}{15.9} \\
\textcolor{black!55}{\quad \texttt{\scriptsize SF\_080\_0010\_A}} & \textcolor{black!55}{86} & \textcolor{black!55}{9.10} & \textcolor{black!55}{3.20} & \textcolor{black!55}{3.28} & \textcolor{black!55}{14.9} & \textcolor{black!55}{3.11} & \textcolor{black!55}{3.10} & \textcolor{black!55}{7.72} & \textcolor{black!55}{2.68} & \textcolor{black!55}{2.72} \\
\textcolor{black!55}{\quad \texttt{\scriptsize SF\_110\_0180\_A}} & \textcolor{black!55}{40} & \textcolor{black!55}{1.41} & \textcolor{black!55}{19.8} & \textcolor{black!55}{22.1} & \textcolor{black!55}{1.17} & \textcolor{black!55}{1.29} & \textcolor{black!55}{1.25} & \textcolor{black!55}{2.35} & \textcolor{black!55}{1.43} & \textcolor{black!55}{4.02} \\
\midrule
\textbf{All scenes} & \textbf{4275} & \textbf{20.9} & \textbf{5.02} & \textbf{5.02} & \textbf{36.6} & \textbf{4.38} & \textbf{4.38} & \textbf{12.9} & \textbf{3.89} & \textbf{4.00} \\
\bottomrule
\end{tabular}

    \caption{\textbf{Source- and scene-level results on ZDPShift benchmark, with video matchers ($\Delta=+32$).}}
    \label{tab:scene_video}
\end{table}